\documentclass{article}
\usepackage{naturetex}

\begin{document}
\captionsetup[figure]{name={Fig.},labelsep=space,labelfont=bf}
\captionsetup[table]{name={Table},labelsep=space,labelfont=bf}
\maketitle

{\setstretch{1.0}
\section*{Abstract}
The working mechanisms of complex natural systems tend to abide by concise and profound partial differential equations (PDEs). Methods that directly mine equations from data are called PDE discovery, which reveals consistent physical laws and facilitates our adaptive interaction with the natural world. In this paper, an enhanced deep reinforcement-learning framework is proposed to uncover symbolically concise open-form PDEs with little prior knowledge. Particularly, based on a symbol library of basic operators and operands, a structure-aware recurrent neural network agent is designed and seamlessly combined with the sparse regression method to generate concise and open-form PDE expressions. All of the generated PDEs are evaluated by a meticulously designed reward function by balancing fitness to data and parsimony, and updated by the model-based reinforcement learning in an efficient way. Customized constraints and regulations are formulated to guarantee the rationality of PDEs in terms of physics and mathematics. The experiments demonstrate that our framework is capable of mining open-form governing equations of several dynamic systems, even with compound equation terms, fractional structure, and high-order derivatives, with excellent efficiency. Without the need for prior knowledge, this method shows great potential for knowledge discovery in more complicated circumstances with exceptional efficiency and scalability.  
}
\\
\noindent\textbf{Keywords:} PDE discovery; symbolic representation; deep reinforcement learning; structure-aware LSTM agent.

\newpage

The laws of physics reveal how the world around us works. Many phenomena in dynamic natural systems can be described by concise and elegant governing equations. The exploration of these equations can be done in two ways: (1) induction, which involves deriving equations directly from data (e.g., Kepler’s laws were built upon Brahe’s observations, as shown in Fig.~\ref{fig:whole_framework}a); and (2) deduction, which involves deriving equations with rigorous mathematics from general rules and principles (e.g., Newton discovered the law of universal gravitation by standing upon the shoulder of Kepler). However, as the system’s complexity and nonlinearity increase and the amount of data multiplies, it becomes increasingly challenging to derive governing equations for such a system. The rapid advancements in computer science have enabled artificial intelligence (AI) to assist scientists in uncovering physical laws. AI aids and stimulates scientific research by using its powerful processing capacity to detect patterns and prove conjectures~\cite{nat_AI}, although human scientists are still needed for this process. The salient question is whether AI can autonomously mine the governing equations from data without the need for prior knowledge.

The essence of mining natural laws in dynamic systems is to identify the relationship between the state variables and their derivatives in space and time through observations, so as to extract the governing equations that can satisfy the laws of physics (e.g., conservation laws) \cite{kemeth2022learning}. Sparse regression is an essential and commonly used method to accomplish the differential equation discovery task. SINDy \cite{sindy} first utilized the sparsity-promoting technique to identify the most important function terms in the preset library that conform to the data, in order to obtain an accurate and concise equation representation of dynamic systems of ordinary differential equations (ODE). After that, SINDy's variants have extended this method to more challenging scenes. PDE-FIND \cite{pde_find} further explored other more complex and high-dimensional dynamic systems described by PDEs, such as the Navier-Stokes equation, using  the method of sequential threshold ridge regression (STRidge). Indeed, state-of-the-art (SOTA) performance was achieved on various problems, such as boundary value problems~\cite{sindy_bvp}, and low-data and high-noise problems~\cite{sindy_pi,weak_sindy,ensemble_sindy,pinn_pdedis}. Despite the remarkable success achieved, this series of methods is limited to a closed and overcomplete candidate library. On the one hand, the selection of candidate functions requires strong prior knowledge; otherwise, the computational burden will be dramatically increased. On the other hand, although sparse regression can determine the possible function terms and their coefficients simultaneously, it can only generate linear combinations of these candidates, and the expressive ability is highly limited.

In order to solve this problem, the expandable library method and symbolic representation method were proposed. PDE-Net series \cite{pde_net,pde_net2} generated new interaction terms based on the topology of the proposed symbolic neural network. Genetic algorithm (GA) expanded the original candidate set through the recombination of gene fragments~\cite{dlga}. Compared with SINDy, they were capable of generating interactive function terms with multiplication and addition operations incorporated. However, it is still deficient in producing the division operation and compound derivatives, much less discovering open-form equations. SGA \cite{sga_pde} further adopted symbolic representations and GA, and represented each function term with a tree structure. Any PDE can be formulated based on the interaction and combination of different function terms. This method markedly increases representation flexibility, but crossover and mutation operations may lead to poor iterative stability of the generated equation form, which then results in a significant increase in computation time. 

Uncovering physical laws through the free combinations of operators and symbols has achieved great progress in symbolic regression due to its great flexibility and few requirements for prior knowledge. The core of this kind of method is mainly divided into two parts: representation and optimization~\cite{chen_inte}. Equations are firstly reasonably represented by the predefined symbols, and then iteratively updated by the optimization algorithms until a desired level of accuracy is reached. Due to the advantages of the evolutionary methods in solving optimization problems, many symbolic regression algorithms based on GA and tree-like representation of structures were put forward \cite{schmidt,gp_sr,gp_sr3,gp_sr2,gp_sr4}.
GA greatly expands the flexibility of representation, but simultaneously causes inefficient optimization and noise sensitivity. With the progress and development of deep learning, neural networks are gradually leveraged to reveal common laws in nonlinear dynamic systems. Their application can be divided into two categories. One is to generate a flexible combination of operations and state variables by adjusting the network topology. The supervised learning task is established to update the neural network  \cite{pde_net2,eql,eql_divide}. The other is based on the idea of reinforcement learning (RL). The initial mathematical expressions are generated by the agent and only expressions with larger rewards are retained to update the agent, which in turn promotes better expressions \cite{sun2022symbolic,dsr}. Compared with GA, gradient-based methods in deep learning show great superiority in searching efficiency but are also insufficient in the diversity of representation. Recently, some attempts to combine GA and RL also indicate a great potential to solve real-world regression problems \cite{rl_gep, dsr_gp}.  Few studies, however, focus on the PDE discovery task. Methods designed for the symbolic regression task, such as DSR \cite{dsr}, focus on identifying a single regression target, and cannot be used to discover governing equations containing abundant physical information (including complex partial differential terms). Due to the focus on fitting the expressions to the measurements, relevant algorithms are liable to suffer from overfitting problems caused by noisy data and generate redundant terms.

To ameliorate the limitation of fixed candidate libraries in the past PDE discovery methods and accelerate the search process, we propose a framework, \textbf{D}eep \textbf{I}dentification of \textbf{S}ymbolically \textbf{C}oncise \textbf{O}pen-form PDEs \textbf{V}ia \textbf{E}nhanced \textbf{R}einforcement-learning (DISCOVER), which can efficiently uncover the parsimonious and meaningful governing equations underlying complex nonlinear systems. Specifically, we design a novel structure-aware long short-term memory (LSTM) agent to generate symbolic PDE expressions. It combines structured input and monotonic attention to make full use of historical and structural information. Compared with random generation, equations generated in an autoregressive manner are more consistent with physical laws and easier to optimize with higher efficiency. STRidge is also well combined to determine the coefficients of function terms. To avoid generating unreasonable expressions and reduce the search space, customized constraints and regulations are designed according to the domain knowledge of physical and mathematical laws. In the evaluation stage, a new reward function is introduced to ensure parsimony of the discovered equations under the condition that it strictly conforms to observations.

Conceptually, this framework seamlessly combines the flexibility of symbolic mathematical representation with the efficiency of reinforcement learning optimization. We demonstrate that  the above framework can handle the discovery of concise PDEs in open form according to experiments on multiple canonical dynamic systems and nonlinear systems in the phase separation field. Its flexibility and computational efficiency have been significantly improved compared to sparse regression methods and GA-based methods, respectively. PDEs even with fractional structure and compound high-order function terms can be well identified.

\section*{Results}

\begin{figure}[!ht]
\centering
\includegraphics[width=\linewidth]{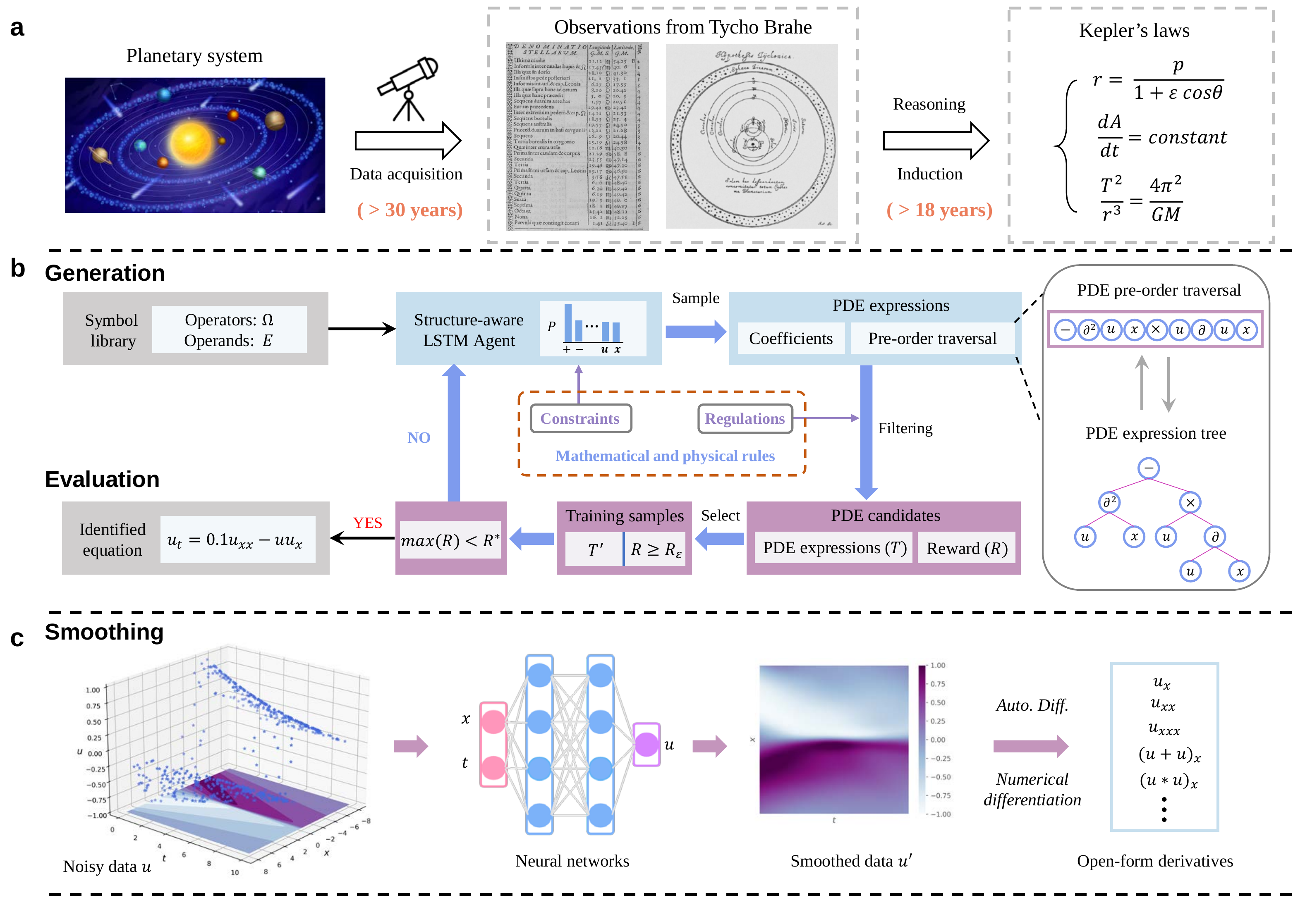}
\caption{\textbf{Illustration of the traditional discovery process and DISCOVER.} \textbf{a} Discovery of Kepler's laws. \textbf{b} and \textbf{c} Overview of the DISCOVER framework. \textbf{b} The main working process including generation and evaluation. The symbol library is composed of operators ($\Omega=\{+,-,\times,\div,\land^2,\land^3,\partial,\partial^2,\partial^3,sin,cos,log\}$ and operands ($E=\{u,x,t,const\}$). The generated PDE traversal sequence can be represented by a unique corresponding PDE tree. $R_{\varepsilon}$ is the $(1-\varepsilon)$-quantile of the rewards of a batch and $R^*$ is the predefined reward threshold to terminate the search process. $T$ refers to the generated PDE expression set that conforms to the laws of physics and mathematics, while ${T}'$ represents the PDE expressions whose rewards are greater than the threshold $R_{\varepsilon}$, which are also the final training samples utilized to update the agent. \textbf{c} Smoothing procedures (preprocessing part), in which the fully connected neural network is utilized to fit the noisy and sparse observations, so as to smooth data and generate metadata on the rest of the spatial-temporal domain. Numerical differentiation or automatic differentiation is incorporated to evaluate the open-form derivatives.}
\label{fig:whole_framework}
\end{figure}

\begin{figure}[!t]
\centering
\includegraphics[width=\textwidth]{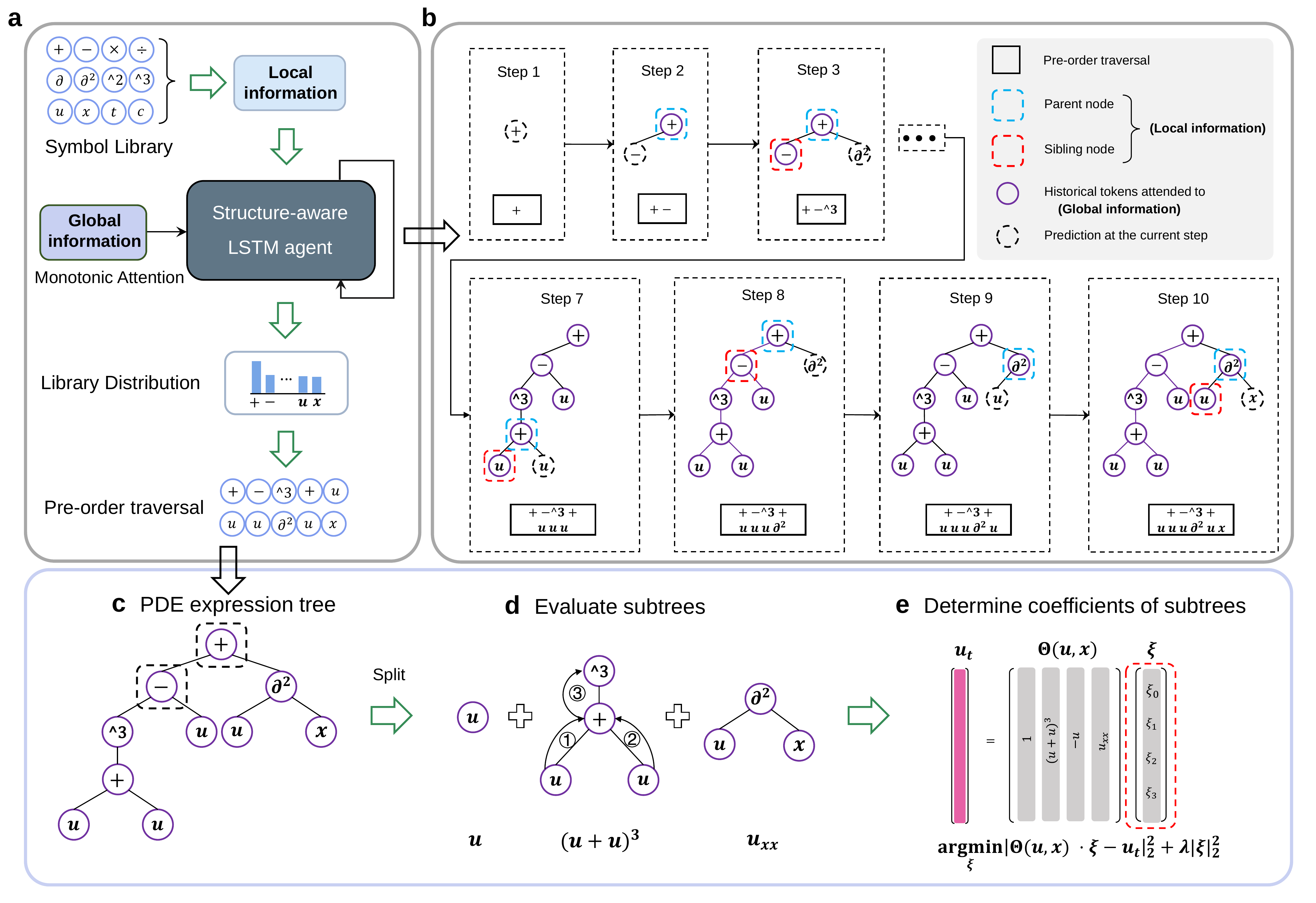}
\caption{\textbf{Example of generating a candidate expression $u_t=u_{xx}-u+0.1250\left(u+u\right)^3$ for the Chafee-Infante equation.} \textbf{a} Procedures of generating pre-order traversal of PDE expression trees by the structure-aware LSTM agent based on symbolic representation. The pre-defined symbol library consists of all of the operators and operands to generate the open-form PDEs. Local information and global information are both incorporated during the generation process. The former serves as inputs to the agent explicitly and the latter is utilized by means of monotonic attention in the form of latent variables of aggregated historical information. The output of the LSTM at each time step is the probability distribution of all tokens in the library, and then one of the tokens is sampled based on it. The pre-order traversal sequence is produced aggressively until the terminal condition is satisfied. \textbf{b} The evolution process of local information and global information utilized during the generation process. Local information includes the parent node and sibling node of the current token in the corresponding expression tree. Global information takes historical information in the past into account. \textbf{c} The PDE expression tree is reconstructed from the corresponding pre-order traversal.  \textbf{d} Split the expression tree into subtrees according to ($+,-$) operators at the top of the tree, and traverse to calculate each term’s value.  \textbf{e} Calculate the coefficients of the function terms based on STRidge.}
\label{fig:generation}
\end{figure}

A nonlinear dynamic system can usually be represented by a parameterized PDE given by:
\begin{equation}
u_{t}=F\left(u, u_{x}, u_{x x}, \cdots, x, \xi\right)=\Theta(u, x) \cdot \xi
\end{equation}
where $u$ is the observations of interest collected from experiments or nature; $u_t$ is the first-order time derivative term; and $F$ is a nonlinear function on the right-hand side, consisting of $u$ and its space derivatives with different orders (e.g., $u_x$ and $u_{xx}$). The coefficients of those candidate terms in $F$ can be represented by $\xi$. $\Theta(u,x)$ is our target with a more concise form.

The overview of DISCOVER uncovering PDEs from data is demonstrated in Fig.~\ref{fig:whole_framework}b, and comprises two parts: generation and evaluation. 
The main purpose of the generation process is to first set up a reasonable symbol library and rules to represent PDEs, and then generate PDE expressions through the agent, including the pre-order traversal of PDEs with a tree structure and coefficients of function terms. The former ones are produced autoregressively by sampling the probability distribution of the agent's output, which is constrained to ensure its rationality in physical and mathematical laws. The latter ones are solved based on the structure of PDE trees and the sparsity-promoting method. During the evaluation stage, the meticulously designed reward function is utilized to evaluate the generated PDE candidates that comply with the customized regulations. Then, the PDE candidates with higher rewards are selected as the training samples. The agent is iteratively updated with the risk-seeking policy gradient method to generate better-fitting expressions until the termination condition is met. Note that observations collected from experiments may be sparse and noisy in real application scenarios. As shown in Fig.~\ref{fig:whole_framework}c, the fully connected neural networks are built to fit and smooth the available data and the metadata are generated to assist the evaluation of derivatives~\cite{DL_PDE} and noise resistance. By introducing this preprocessing part, in essence, the neural network (NN) in DISCOVER is built as a surrogate model to learn the mapping relationship between spatial-temporal inputs and states of interest. It is then analyzed for interpretability and converted into a human-readable format (PDE). In aggregate, DISCOVER can identify open-form governing equations directly from the data, while also possessing high efficiency and scalability. The high efficiency is due to three innovations of our framework: (1) DISCOVER utilizes a structure-aware agent to produce PDE expressions with a tree structure; (2) the coefficients of PDEs are determined by deconstructing the equation tree  and combining the sparsity-promoting method to avoid multiple constant optimization processes; and (3) the whole optimization process of generating PDE expressions is neural-guided and parallelizable. In terms of scalability, the symbol library can be easily expanded, and customized constraints and regulations can be incorporated to accelerate the search process according to prior knowledge. Additional details about the frameworks can be found in Methods and Supplementary Information. We will demonstrate the superiority of DISCOVER through canonical problems.

\subsubsection*{Discovering open-form PDEs of canonical dynamic systems.}
We verified the accuracy and efficiency of DISCOVER in mining open-form PDEs containing strong nonlinearities with little prior knowledge. Specifically, we utilized the study cases from previous studies~\cite{sga_pde, pde_find,ensemble_sindy, EPDE}, including Burgers’ equation with nonlinear interaction terms, the KdV equation which has high-order derivatives, the Chafee-Infante equation with exponential terms, and the viscous gravity current equation (PDE\textunderscore compound) with compound function terms and fractional structure (PDE\textunderscore divide). We use these to test its performance and compare it against a GA-based model (SGA~\cite{sga_pde}), which is the latest PDE discovery model to handle complex open-form equations. The results show that our framework is not only accurate but also computationally efficient and stable.

\paragraph{Study cases.} Five canonical models of mathematical physics include: (1) \textbf{The KdV equation:} It was jointly discovered by Dutch mathematicians Korteweg and De Vries to describe the one-way wave on shallow water \cite{kdv}. It takes the form of $u_t=auu_x+bu_{xxx}$, where $a$ is set to -1, and $b$ is set to -0.0025. 
(2) \textbf{Burgers’ equation:}
As a nonlinear partial differential equation that simulates the propagation and reflection of shock waves, Burgers’ equation is widely used in numerous fields, such as fluid mechanics, nonlinear acoustics, gas dynamics, etc.\cite{burgers}. We consider a 1D viscous Burgers’ equation $u_t=-uu_x+vu_{xx}$, where $v$ is equal to 0.1.
(3) \textbf{The Chafee-Infante equation:}
Another 1D nonlinear system is $u_t=u_{xx}+a\left(u-u^3\right)$ with $a=-1$, which was developed by Chafee and Infante \cite{chafee_1}. It has been broadly employed in fluid mechanics~\cite{chafee1}, high-energy physical processes~\cite{chafee2}, electronics~\cite{chafee3}, and environmental research \cite{chafee_2}. 
(4) \textbf{PDE\textunderscore compound} and \textbf{PDE\textunderscore divide:}
These two equations are first put forward in SGA and used to demonstrate that our framework is capable of mining open-form PDEs, which is crucial for extracting unidentified and undiscovered governing equations from observations. PDE\textunderscore compound with equation form $u_t=(uu_x)_x$ is proposed to describe viscous gravity currents with a compound function \cite{pde_compound}. PDE\textunderscore divide with equation form $u_t=-u_x/x+0.25u_{xx}$ contains the fractional structure that conventional closed-library-based methods cannot handle. 

\begin{figure}[!ht]
\includegraphics[width=\textwidth]{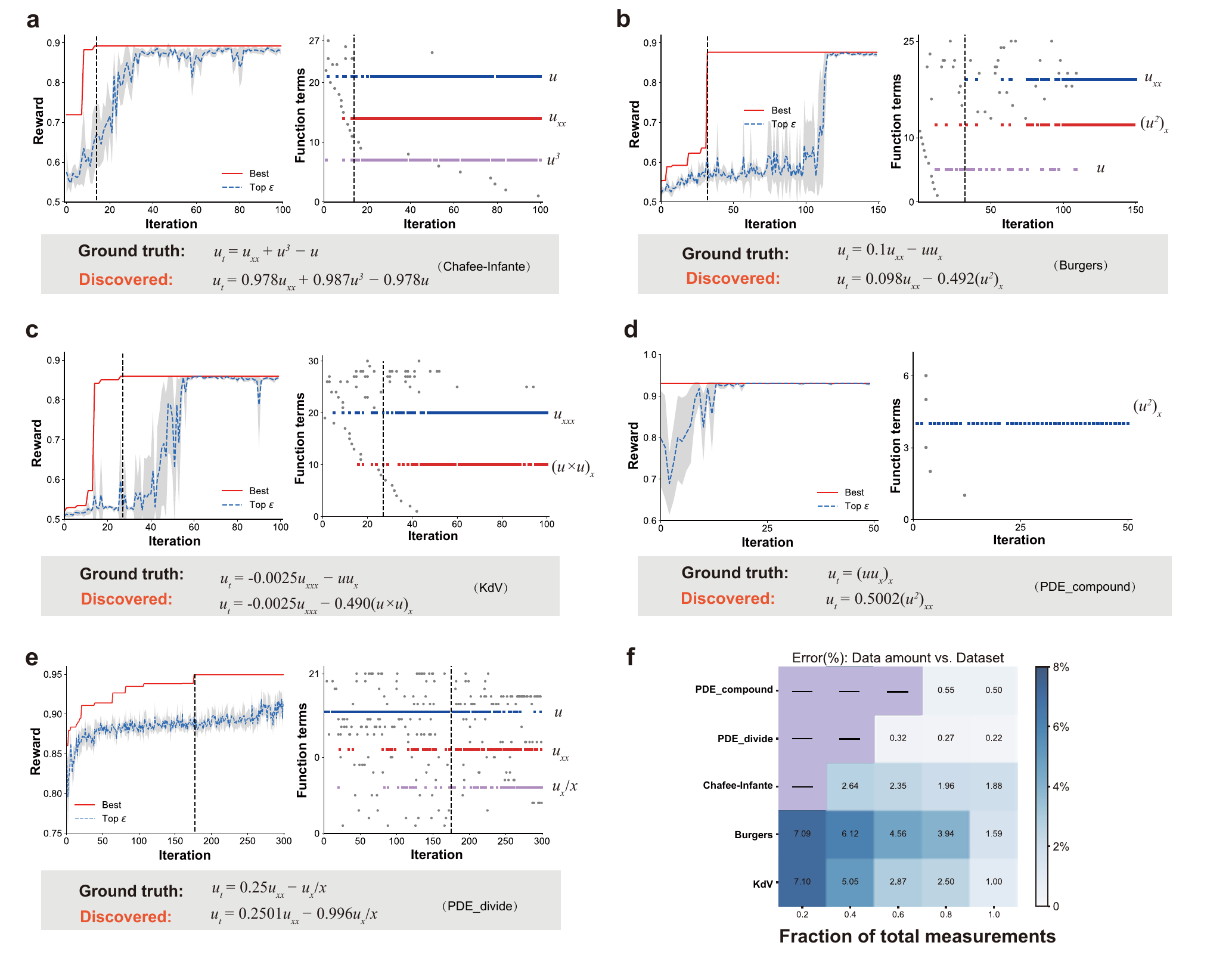}
\caption{ 
\textbf{Discovery process of some canonical physical problems with noisy and sparse measurements.} \textbf{a-e}: The evolution of rewards and function terms of the best-case expression of \textbf{a} the Chafee-Infante equation, \textbf{b} Burgers' equation, \textbf{c} the KdV equation, \textbf{d} PDE\_compound and \textbf{e} PDE\_divide. 10\% noise in the first three equations is used, and 1\% noise is used in the last two equations. The figure on the left denotes the evolution of the best reward so far and the reward of the top $\varepsilon$ expressions (utilized to update the agent). The figure on the right denotes the evolution of function terms of the best case in the current iteration. Function terms that occur the most are individually marked with colored dots. The gray dashed-line marks the iteration where the expression with maximum reward appears. \textbf{f} Relative error of different physical systems with different volumes of data. Squares marked with a solid black line indicate that the correct form of the equation could not be found.}
\label{fig:noise}
\end{figure}

The default hyperparameters used to mine the above PDEs are provided in section S1.5 of Supplementary Information. Table~\ref{results} shows that under the premise of little prior knowledge, DISCOVER can uncover the analytic representation of the various physical dynamics mentioned above, and the discovered equation form is accurate and concise even with Gaussian noise added. 
Fig.~\ref{fig:noise} shows the performance of DISCOVER in the mining equation under the condition of noisy and sparse data. Fig.~\ref{fig:noise} a-e illustrates the evolution of rewards and function terms of the best expression in the optimization process. For complex equations, there will be redundant or incorrect terms in the generated expressions during iteration (e.g., function term $u$ in Burgers and PDE\_divide). However, since model-based optimization is directional, more accurate expressions will be generated as rewards of better cases increase and fewer redundant terms appear. In addition, a risk-seeking policy gradient is utilized and DISCOVER is able to find the optimal result with only the best-case performance considered. Meanwhile,  more available data collected contributes to a more accurate surrogate model, and then it is more likely to uncover the correct equation, as shown in Fig.~\ref{fig:noise}f. More discussion on noise and data sparsity can be found in section S2.3 of Supplementary Information. It is worth noting that the KdV equation, Burgers’ equation, and the Chafee-Infante equation can be discovered correctly by conventional sparse regression methods and genetic algorithms, but not for the PDE\textunderscore compound and PDE\textunderscore divide equations. This is primarily because these methods rely on a limited set of candidates, and cannot handle equations that contain compound terms or fractional structures. The result demonstrates that our proposed method can directly mine open-form equations from data, which offers wider application scenarios and practicability.

\begin{table}[t]
\setlength{\abovecaptionskip}{0cm} 
\setlength{\belowcaptionskip}{0.2cm}
\caption{\textbf{Summary of discovered results for different PDEs of mathematical physics.} The subscripts $m$ and $n$ denote the number of discretizations. For noisy measurements, 10\% noise is added for the KdV, Burgers', and Chafee-Infante equations, and 1\% noise is added for PDE\_divide and PDE\_compound.}
  \label{results}
  \centering
    \sffamily
 \renewcommand{\arraystretch}{1.5}
 \resizebox{\textwidth}{!}{
\begin{tabular}{lccc}
\toprule
\multicolumn{1}{c}{\large \textbf{PDE systems}} & \large \textbf{PDE discovered (clean data)}  & \large \textbf{Error (clean, noisy)} & \large  \textbf{Data discretization}         \\ \midrule\addlinespace[1ex]
    \begin{minipage}{.2\textwidth}
      \includegraphics[width=\linewidth, height=18mm]{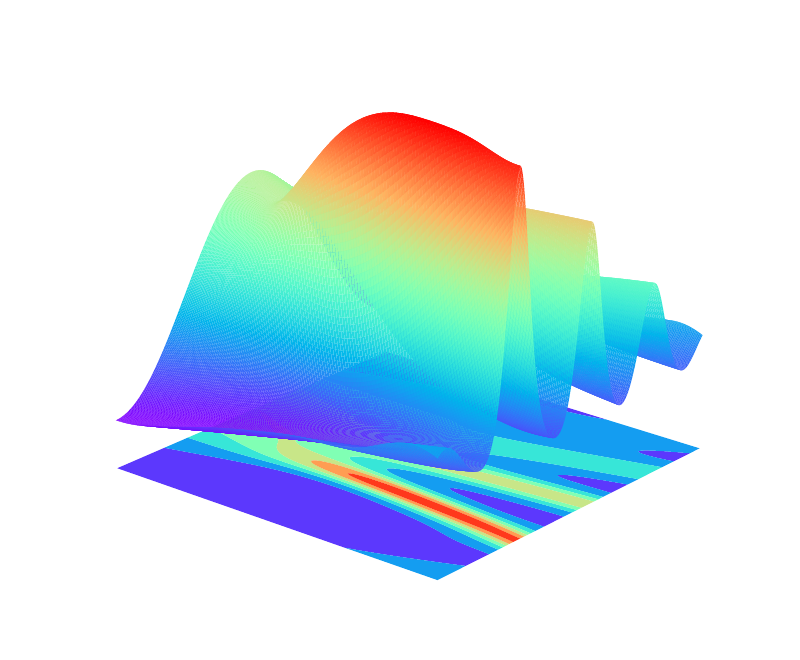}
    \end{minipage} KdV
 & $u_t=-0.5001(u\times u)_x-0.0025u_{xxx}$             & $0.09\pm0.07\%$, $1.0\pm1.41\%$ &    $x \in[-1,1)_{m
=512}, t \in[0,1]_{n=201}$            \\
    \begin{minipage}{.2\textwidth}
      \includegraphics[width=\linewidth, height=20mm]{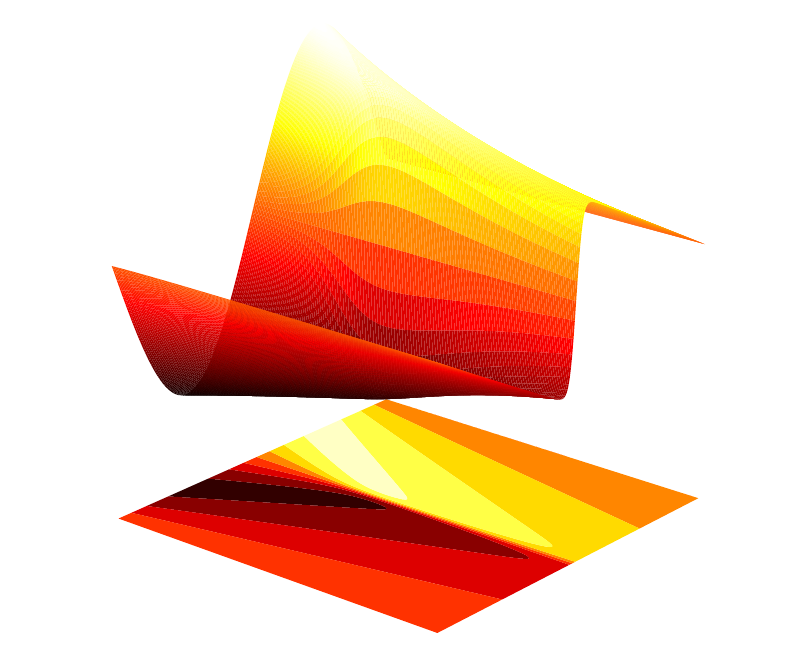}
    \end{minipage}  Burgers  
&$u_t=-1.0010uu_x+0.1024u_{xx}$         &$1.34\pm1.61\%$,$1.59\pm0.41\%$ & $x \in[-8,8)_{m
=256}, t \in[0,10]_{n=201}$               \\
    \begin{minipage}{.2\textwidth}
      \includegraphics[width=\linewidth, height=20mm]{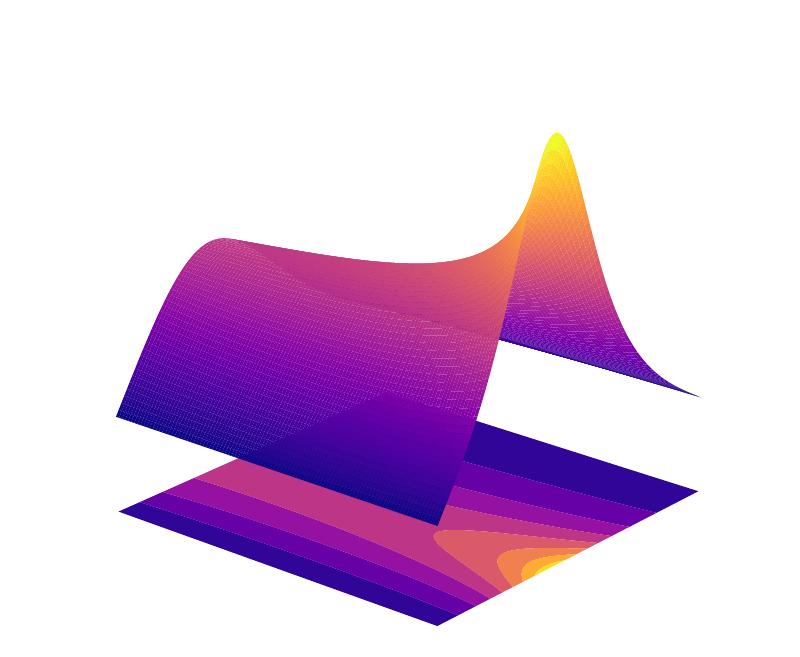}
    \end{minipage} Chafee-Infante
&$u_t=1.0002u_{xx}-1.0008u+1.0004u^3$            &$0.04\pm0.0\%$,$1.88\pm0.54\%$ &  $x \in[0,3]_{m
=301}, t \in[0,0.5]_{n=200}$       \\
    \begin{minipage}{.2\textwidth}
      \includegraphics[width=\linewidth, height=20mm]{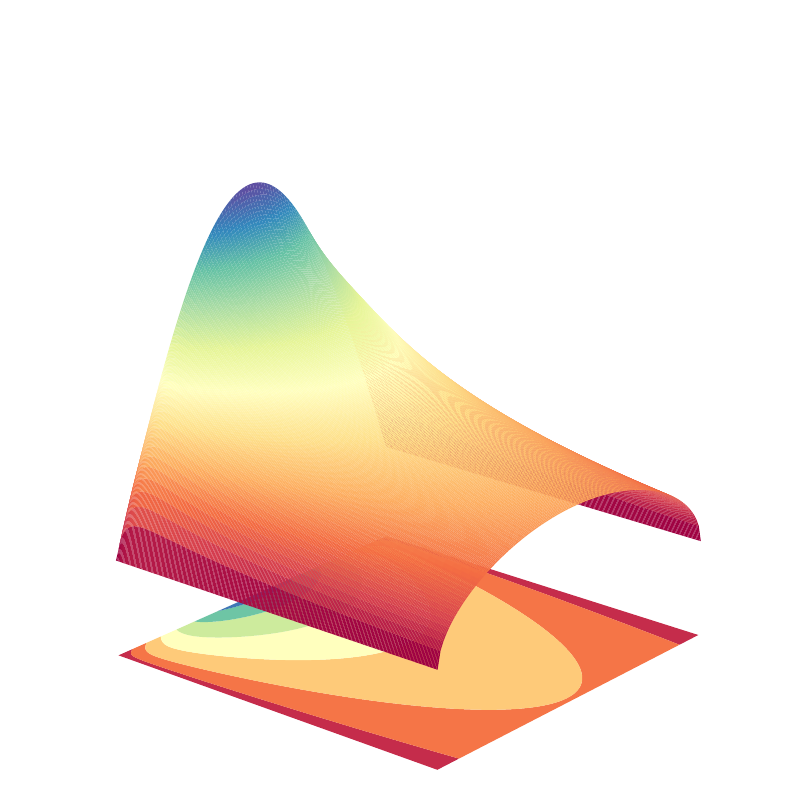}
    \end{minipage}  PDE\textunderscore compound 
& $u_t=0.5002(u^2)_{xx}$  & $0.04\pm0\%$, $0.05\pm10\%$        & $x \in[1,2)_{m
=301}, t \in[0,0.5]_{n=251}$    \\
    \begin{minipage}{.2\textwidth}
      \includegraphics[width=\linewidth, height=20mm]{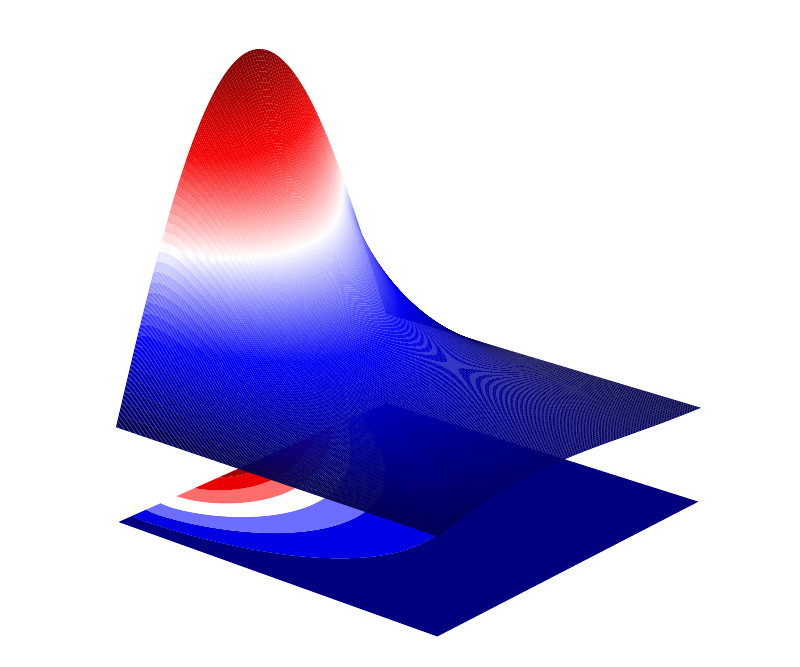}
    \end{minipage} PDE\textunderscore divide
&  $u_t=-0.9979u_x/x+0.2498u_{xx}$            & $0.14\pm0.10\%$, $0.22\pm0.25\%$   &$x \in[1,2)_{m
=100}, t \in[0,1]_{n=251}$\\ \bottomrule

\end{tabular}
}
\end{table}

\subsubsection*{Comparisons to the GA-based method.}
At the expense of flexibility and representation ability, common PDE discovery methods such as SINDy utilize a stationary library and sparse regression to uncover physical laws. Since the stationary library cannot cover or generate all of the complex structures, they are deficient in identifying complex equation forms such as PDEs with fractional structures. For fairness of comparison, the latest symbolic PDE discovery method SGA is taken as an example, which is also capable of uncovering the equation representations of these five physical dynamics correctly. Specifically, we will compare the specific performance (time consumption and accuracy) of DISCOVER based on RL and SGA based on GA. Note that all of the experiments were replicated with five different random seeds for each PDE. As shown in Table~\ref{com_sga}, the left column represents the MSE between the left-hand and right-hand sides of the discovered equations. A smaller MSE means a more accurate discovered PDE. Both DISCOVER and SGA are capable of mining all terms of the equation exactly, but their coefficients of function terms are slightly dissimilar, resulting in the former having slightly smaller errors than the latter in the first four equations. The right column of Table~\ref{com_sga} presents the time consumed by the two methods to identify the optimal equation. It can be seen that, in addition to the Chafee-Infante equation, our method uses only approximately 40\% of the time cost by SGA(fast) (a more efficient version) on average, which has higher computational efficiency. This is mainly because our method is model-based, and the entire optimization process is directional with a positive gain. As the iteration proceeds, the generated equations become increasingly reminiscent of the authentic expression. Additional details of the optimization process can be found in section S2.2 of Supplementary Information. In contrast, SGA expands the diversity of the generated equation representations primarily through crossover and mutation of gene fragments, which is more stochastic and uncontrollable. Although it facilitates the search for more complex equations, it also leads to more requisite computational time.

It is worth noting that in the process of uncovering the Chafee-Infante equation, SGA takes $u$ as a default function term (this information itself is known and easily accessible). By introducing this prior knowledge, the optimal form of the equation can always be easily found in the first round of iterations. Without this knowledge, however, SGA is unable to identify the correct form of the equation within 300 iterations (>1000 s). The main reason for this problem is that SGA is modeled based on function terms that are randomly generated, and each function term is represented by a multi-layer tree structure, while the term $u$ is represented as a one-layer root node. In other words, the way that function terms are defined makes SGA prone to produce complex tree structures. However, DISCOVER models the equation as a whole with a pre-order traversal sequence and then partitions it according to the operators ("+" or "-"). The symbol sequence is autoregressively generated and consequently, both simple and complex function terms can be handled easily. It can be seen that our method is time-efficient and has better practicality with little prior knowledge.

\begin{table}[!ht]
\setlength{\abovecaptionskip}{0cm} 
\setlength{\belowcaptionskip}{0.2cm}
\small
 \caption{\textbf{Comparison of MSE of discovered results and running time for DISCOVER and SGA.}}
\begin{threeparttable}[t]
\centering
\renewcommand{\arraystretch}{1.5}
\resizebox{\textwidth}{!}{
\begin{tabular}{ccccccc}
\hline
\rowcolor{Gray}
\multirow{2}{*}{} & \multicolumn{2}{c}{\textbf{MSE}} & \multicolumn{4}{c}{\textbf{Runing Time (s)}} \\ \cline{2-7} 
    &DISCOVER            & SGA                    & DISCOVER  & SGA  & SGA(fast)\tnote{1} & SGA(w/o $u$)\tnote{2} \\ \hline 
KdV   & \bm{$3.16\times10^{-5}$}              & $1.48\times10^{-4}$              & \bm{$243.66$}   & 1464.80  & 890.50       & \textbackslash{}              \\\rowcolor{Gray}
Burgers                        & \bm{$4.89\times10^{-7}$}              & $4.33\times10^{-5}$              & \bm{$206.76$}   & 495.18 & 423.8      & \textbackslash{}  \\
Chafee-Infante                 & \bm{$1.36\times10^{-4}$}              & $4.72\times10^{-4}$              & 67.23    & 27.70  & \bm{$20.12$}            & \textgreater{1000}       \\\rowcolor{Gray}
PDE\_compound                              & \bm{$8.31\times10^{-6}$}               & $4.56\times10^{-5}$              & \bm{$13.31$} & 604.10 &557.20          & \textbackslash{} \\
PDE\_divide                                & $7.64\times10^{-4}$               & \bm{$1.78\times10^{-4}$}            & \bm{$1259.53$}  & 2046.24 & 1466.51      & \textbackslash{}   \\\hline
\end{tabular}
}
\begin{tablenotes}
 \item[1] SGA(fast) is an optimized version of the original article code, with a faster computation speed.
 \item[2] SGA(w/o $u$) represents the SGA(fast) version without prior knowledge $u$ provided in advance.
\end{tablenotes}
\end{threeparttable}

\label{com_sga}
\end{table}

\begin{figure}[!ht]
\includegraphics[width=\textwidth]{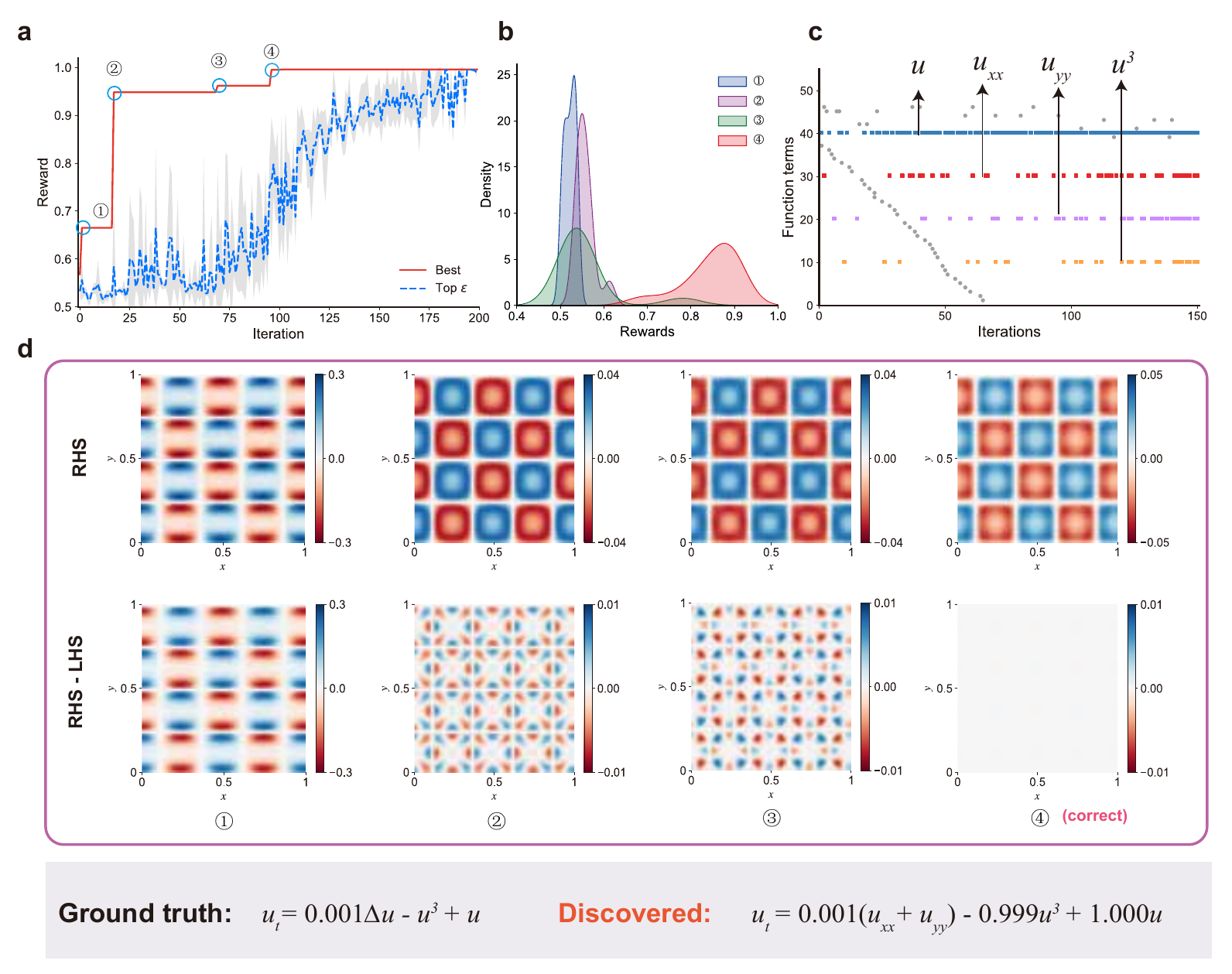}
\caption{ 
\textbf{The optimization process and intermediate results at the $40^{th}$ time step of discovering the Allen-Cahn equation.} \textbf{a} The distribution of the best-case reward and top $\varepsilon$ fraction of rewards. \normalsize{\textcircled{\scriptsize{1}}}\normalsize-\normalsize{\textcircled{\scriptsize{4}}}\normalsize\, represents the positions where the reward goes through a sharp jump. \textbf{b} The Gaussian kernel density estimate of the top $\varepsilon$ fraction of rewards at \normalsize{\textcircled{\scriptsize{1}}}\normalsize-\normalsize{\textcircled{\scriptsize{4}}}\normalsize. \textbf{c} Evolution of the function terms of the best expression. The Y-axis corresponds to the 90 most frequent equation terms with decreasing occurrences from the top to the bottom. Each point represents the corresponding function term appearing in the best-case expression produced by the agent at the current iteration step. The four most frequently occurring equation terms are $u$, $u_{xx}$, $u_{yy}$, and $u^3$, respectively, which are taken out separately and correspond to function terms of the correct equation form. \textbf{d} The change of the RHS of the mined PDE and the difference between the RHS and LHS ($u_t$), which is the residual of the equation.}
\label{fig:al_ch equation}
\end{figure}

\subsubsection*{Discovering complex open-form PDEs of phase separation.}
To demonstrate that the proposed DISCOVER framework is capable of mining the high-dimensional PDEs with high-order derivatives, we introduce the Allen-Cahn and Cahn-Hilliard equations, which are firmly nonlinear gradient flow systems and frequently used to describe phase separation processes in fluid dynamics~\cite{ac_ch_1, ac_ch_2, ch_1} and material sciences\cite{ac_ch_phase, ch_2}.

\paragraph{The Allen-Cahn equation.} 
The 2D reaction-diffusion systems considered here can be represented by 
$u_t=\epsilon^{2}\Delta u - f(u)$, where $\epsilon$ is a small constant value representing the interfacial thickness and $\epsilon^{2}$ is set to 0.001; $\Delta u$ denotes the diffusion term; and $f(u) = u^3-u$ represents the reaction term. For this example, Laplace operators appear in the equation in addition to the increase in spatial dimensionality. To ensure that the PDEs mined by DISCOVER conform to the physical laws, we set a constraint on the spatial symmetry of the generated equations that the order of the partial derivatives with respect to the spatial inputs $x$ and $y$ must be the same for the right-hand-side (RHS) term of the equation, i.e., the number of occurrences of $x$ and $y$ in the traversal sequence is the same. Fig.~\ref{fig:al_ch equation}a shows the trend of best-case reward and top $\epsilon$ rewards. It can be seen that the best-case reward soon reaches 0.95, and as the iteration proceeds, the reward makes four large jumps in total. According to Fig.~\ref{fig:al_ch equation}b the rewards of better-fitting expressions gradually increase at those four jumps. Their discovered equation gradually approaches the ground truth which is demonstrated by Fig.~\ref{fig:al_ch equation}d, where the distribution of the RHS of the equations approaches the true distribution, and the gap between the left-hand-side (LHS) term $u_t$ and it (i.e., the residual) decreases and finally approaches zero. Fig.~\ref{fig:al_ch equation}c illustrates the evolution of the equation terms of the best-case expression during the iterations. As the optimization process proceeds, the gradual increase in the number of equation terms included in the correct equation form proves that the expressions generated by the agent are becoming increasingly accurate while other incorrect function terms gradually disappear. 

\paragraph{The Cahn-Hilliard equation.} 

\begin{figure}[!t]
\includegraphics[width=\linewidth]{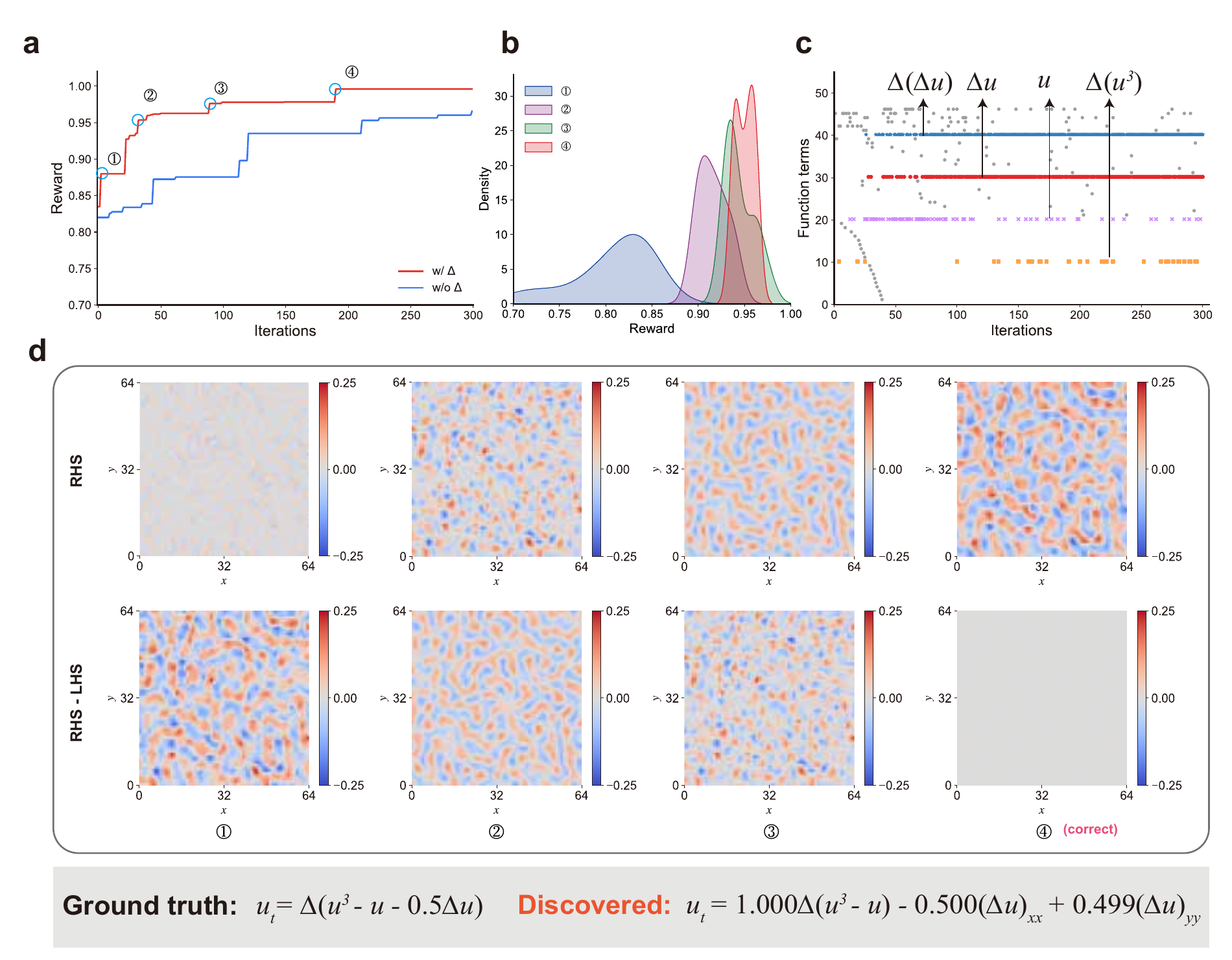}
\caption{
\textbf{The optimization process and intermediate results at the $250^{th}$ time step of discovering the Cahn-Hilliard equation.} \textbf{a} The distribution of the best-case reward and top $\varepsilon$ fraction of rewards. \normalsize{\textcircled{\scriptsize{1}}}\normalsize-\normalsize{\textcircled{\scriptsize{4}}}\normalsize\, represents the positions where the reward goes through a sharp jump. \textbf{b} The Gaussian kernel density estimate of the top $\varepsilon$ fraction of rewards with the Laplace operator at \normalsize{\textcircled{\scriptsize{1}}}\normalsize-\normalsize{\textcircled{\scriptsize{4}}}\normalsize. \textbf{c} Evolution of the function terms of the best expression. The Y-axis corresponds to the 100 most frequent equation terms with decreasing occurrences from the top to the bottom.  The four most frequently occurring equation terms are $\Delta(\Delta u)$, $\Delta u$, $u$, and $\Delta (u^3)$, respectively. \textbf{d} The change of the RHS of the mined PDE and the difference between the RHS and LHS ($u_t$), which is the residual of the equation.
}
\label{fig:ch_hi equation}
\end{figure}

With the fourth order of derivatives and more complicated compound terms, we consider a specific form of Cahn-Hilliard equation $u_t=\Delta(u^3 - u -\kappa\Delta u)$, where $\kappa=0.5$ denotes the surface tension at the interface.
The internal term $u^3-u-\kappa\Delta u$ within the Laplace operator denotes the chemical potential and has an extremely complicated expansion. For this reason, while retaining the symmetric constraint, we introduce a new Laplace operator in the symbol library and compare it with the default library configuration. It can be seen from Fig.~\ref{fig:ch_hi equation}a that the reward with the Laplace operator grows faster and can uncover the true equation form in $194^{th}$ iterative steps. However, by the default configuration, the correct equation form can be identified in the $903^{rd}$ iterative steps, which takes more than four times as many iterative steps. Fig.~\ref{fig:ch_hi equation}b provides the reward distribution of marked iterations and the evolution of function terms is illustrated in Fig.~\ref{fig:ch_hi equation}c. Note that equation term $u$ mainly appears in the beginning stage of training. With the updating of the agent, the occurrence of redundant term u gradually decreases. The several intermediate results in the evolution process and the values of the RHS term and the residual are also shown in Fig.~\ref{fig:ch_hi equation}d. The results demonstrate that our framework can seamlessly incorporate possible prior knowledge by expanding constraint and regulation systems or the symbol library to facilitate the searching process and uncover the underlying physical laws in dynamic systems with strong nonlinearity.

These two examples illustrate that DISCOVER can handle high-dimensional and high-order dynamical systems and conveniently introduce new constraints and operators with great scalability to accelerate the discovery process.
\paragraph{Comparisons with other PDE discovery methods.}

\begin{table}[tbp]
\setlength{\abovecaptionskip}{0cm} 
\setlength{\belowcaptionskip}{0.2cm}
 \caption{Comparisons of DISCOVER and different methods on the PDE discovery and symbolic regression tasks }
  \label{table:compare_results}
  \centering
    \sffamily
\renewcommand{\arraystretch}{1.5}
\resizebox{\textwidth}{!}{
\begin{tabular}{cccccccccc}
\hline
\rowcolor{Gray}
\large \textbf{Equations} & \large \textbf{Correct expressions} & \begin{tabular}[c]{@{}c@{}} \large \textbf{PDE-FIND}\\~\cite{pde_find}\end{tabular} & \begin{tabular}[c]{@{}c@{}}\large \textbf{PDE-NET}\\~\cite{pde_net,pde_net2}\end{tabular} & \begin{tabular}[c]{@{}c@{}}\large \textbf{ Weak form}\\~\cite{weak_sindy}\end{tabular} &\begin{tabular}[c]{@{}c@{}}\large \textbf{EPDE} \\~\cite{EPDE}\end{tabular} & \begin{tabular}[c]{@{}c@{}}\large \textbf{DLGA} \\~\cite{dlga}\end{tabular} &\begin{tabular}[c]{@{}c@{}}\large \textbf{SGA} \\~\cite{sga_pde}\end{tabular} & \begin{tabular}[c]{@{}c@{}}\large \textbf{DSR}\\~\cite{dsr,dsr_gp}\end{tabular} & \large \textbf{DISCOVER} \\ \hline
KdV   &$u_t=-uu_x-0.0025u_{xxx}$      &   $\surd$      &    $\surd$   &    $\surd$      &   $\surd$       &   $\surd$    &    $\surd$      & \notcheckmark    &  $\surd$   \\ \rowcolor{Gray}
Burgers & $u_t=-uu_x+0.1u_{xx}$       &   $\surd$        &    $\surd$       &    $\surd$  &  $\surd$       &   $\surd$    &    $\surd$      & \notcheckmark    &  $\surd$         \\
Chafee-Infante   &$u_t=u_{xx}+u-u^3$    &   $\surd$    &    $\surd$      &       &   $\surd$       &   $\surd$    &    $\surd$      & \notcheckmark   &  $\surd$          \\\rowcolor{Gray}
PDE\_compound  &$u_t=(uu_x)_x$       &          &          &   \notcheckmark       &         &      &  $\surd$       &   \notcheckmark    &      $\surd$     \\
PDE\_divide  &$u_t=-u_x/x+0.25u_{xx}$       &          &          &           &         &      &      $\surd$   &   \notcheckmark   &        $\surd$   \\\rowcolor{Gray}
Allen-Cahn &  $u_t=0.001\Delta u - u + u^3$      &  $\surd$        &  $\surd$        &  $\surd$           & $\surd$          & $\surd$       &  \notcheckmark       &   \notcheckmark &        $\surd$   \\
Cahn-Hilliard &  $u_t=\Delta(u^3 - u -0.5\Delta u)$    &          &         &  \notcheckmark          &         &      &        \notcheckmark &      \notcheckmark &         $\surd$  \\\rowcolor{Gray}
Nguyen benchmarks~\cite{Nguyen} & Nguyen-(1-12) &      &          &           &         &      &    \notcheckmark &   $\surd$   &         \notcheckmark  \\ \hline
\end{tabular}
}
\end{table}
As shown in Table~\ref{table:compare_results}, we have presented whether other methods can identify the equations from data mentioned in this article and summarized the boundaries of different methods.  Closed library methods including PDE-FIND~\cite{pde_find}, and PDE-NET~\cite{pde_net} and expandable library methods including PDE-NET 2.0~\cite{pde_net2}, EPDE~\cite{EPDE}, and DLGA~\cite{dlga} can find the linear combination of possible candidate function terms in the library and are capable of handling the KdV, Burgers, Chafee-Infante, and Allen-Cahn equations. The difference is that the closed library depends more on prior knowledge to ensure that all of the function terms that may appear in the equation are included in the library. Weak-form methods~\cite{weak_sindy} can partly handle the equations with compound terms, but the number of integrals needs to be determined in advance, and fail to deal with multi-layer compound structures. The methods with symbolic representation including SGA~\cite{sga_pde}, DSR~\cite{dsr,dsr_gp}, and DISCOVER can not only uncover the conventional equations, such as the KdV equation, but also the complex PDEs with compound terms and fractional structures, and symbolic regression tasks. Among them, SGA utilizes GA to expand the search space, but is less efficient and currently confined to 1D dynamics and clean data. DSR and DISCOVER are both based on deep reinforcement learning methods; 
whereas DSR is designed for symbolic regression tasks whose optimization objective is not consistent with PDE discovery tasks. Specifically, the current DSR neither incorporates the computation of partial differential operators nor is it able to handle sparse and noisy data. The coefficients also cannot be determined efficiently, which may lead to prohibitive computational costs for high-dimensional nonlinear dynamic systems with large data volumes. A detailed description can be found in Supplementary Information S2.4. DISCOVER performs better for PDE discovery tasks as it balances diversity and efficiency in the mining equation process with better scalability. Although not specifically for regression tasks, it can solve most of the symbolic tasks in Nguyen benchmarks~\cite{Nguyen} under the condition that the appropriate library is defined.


\section*{Discussion}
We propose a new framework named DISCOVER for exploring open-form PDEs based on enhanced deep reinforcement learning and symbolic representations. It reduces the demand for prior knowledge and is capable of dealing with compound terms and fractional structures that conventional library-based methods cannot handle. The structure-aware architecture proposed is capable of learning the PDE expressions more effectively and can be applied to other problems with structured inputs. Moreover, this framework achieves more efficient and stable performance by means of the neural-guided approach and sparsity-promoting methods compared to GA-based methods (e.g., SGA). In addition to some nonlinear canonical 1D problems (Burgers’ equation with interaction terms, the KdV equation with high-order derivatives, the Chafee-Infante equation with derivative and exponential terms, and PDE\_compound and PDE\_divide with compound terms and fractional structure, respectively), we also demonstrate this framework on high-dimensional systems with high-order derivatives in the governing equations. Results show that the proposed framework is capable of uncovering the true equation form from noisy and sparse data, and can be applied to solving new tasks efficiently without known physical knowledge incorporated.

Results demonstrated that the proposed framework can assist researchers in different fields to comprehend data and further uncover the underlying physical laws effectively. To extend the boundaries of knowledge and serve a wider range of applications, there are two potential improvements that should be considered in future work. Firstly, despite the optimization efficiency being greatly improved by model-based methods, we still need to reasonably preserve the diversity of the generated equation forms to cope with the potential exploration and exploitation dilemma. Moreover, robustness to noise may be further strengthened with discovered physical knowledge incorporated. Combining the automatic machine learning methods~\cite{autoke,baydin2018automatic} may have the potential to enhance DISCOVER to further identify the open-form PDEs from high-noise and sparse data.

\section*{Methods}
The steps of DISCOVER mining PDEs mainly consist of three steps: (1) we first build a symbol library and define that a PDE can be represented as a tree structure. A structure-aware recurrent neural network agent by combining structured inputs and monotonic attention is designed to generate the pre-order traversal of PDE expression trees; (2) the expression trees are then reconstructed and split into function terms, and their coefficients can be calculated by the sparse regression method; and (3) all of the generated PDE candidates are first filtered by physical and mathematical constraints, and then evaluated by a meticulously designed reward function considering the fitness to data and the parsimony of the equation. We adopt the risk-seeking policy gradient to iteratively update the agent to improve the better-fitting expressions until the best-case expression meets the accuracy and parsimony requirements that we set in advance. The first two steps correspond to the Generation part, and the third step introduces the Evaluation part in Fig.~\ref{fig:whole_framework}. Additional details are discussed below.

\subsubsection*{Generating the pre-order traversals of the PDE expression trees.}

\paragraph{PDE expression tree.} As shown in Fig.~\ref{fig:whole_framework}a, a PDE can be represented by a binary tree via symbolic representation, and all of the tokens on the nodes are selected from a pre-defined library $\mathcal{L}$. The library consists of two categories of symbols: operators (the first two rows) and operands (the bottom row). Compared with the symbolic regression problem, our library also introduces a differential operator with different orders to calculate the time and space derivatives of state variables. Note that since the LSTM agent is adopted in DISCOVER, which only produces sequential data step by step, pre-order traversal sequences of PDE expression trees are generated in the Generation part.

For a PDE expression tree, the interior nodes are all operators, the leaf nodes are all operands, and their arities are known. For example, the partial derivative $\partial$ is a binary operator with two children, and the space input $x$ is an operand with zero children. This property ensures that each expression tree has a unique pre-order traversal sequence corresponding to it. As a consequence, we can conveniently generate batches of pre-order traversal sequences by means of the LSTM agent instead of the expression trees. 
An expression tree and its pre-order traversal can be represented as $\tau$. The $i$-th token in the traversal can be represented as $\tau_i$ and corresponds to an action under the current policy in reinforcement learning. When generating the token at the current time step, the output of LSTM will be normalized to generate a probability distribution of all tokens in $\mathcal{L}$. The current action is then sampled based on this distribution. A full binary tree is constructed when all leaf nodes in the expression tree are operands, and the generation process is terminated.

\paragraph{Structure-aware LSTM agent.}
The common LSTM is an autoregressive model, which means that predicting the current token $\tau_i$ is conditioned on the last predicted token $\tau_{i-1}$. Because all of the history information is coupled and stored in a cell, it is insufficient for LSTM to deal with long sequences and structural information. To effectively generate PDE expression trees that have strong structural information, we propose a structure-aware architecture for the LSTM agent. Specifically: (1) the structured input is utilized to convey local information as DSR \cite{dsr}. For example, as shown in the seventh time step in Fig.~\ref{fig:generation}b, the inputs are its parent ($+$) and sibling ($^{\wedge}3$), instead of the previous token $u$; and (2) the monotonic attention is leveraged to endow the agent with a more powerful capacity to capture global dependencies and avoids the loss of long-distance information in the step-by-step transmission process. The details of the monotonic attention layer are provided in Supplementary Information S1.2. 

\paragraph{Constraints and regulations.}

Generating PDE expression sequences in an autoregressive manner without restrictions tends to produce unreasonable samples. To reduce the search space and time consumption, we design a series of constraints and regulations based on mathematical rules and physical laws. Constraints are applied to the agent in its generation of the pre-order traversal sequences and can be divided into two categories, including (1) complexity limits of PDE expressions (e.g., the total length of the sequence shall not exceed the maximum length); (2) relationship limits between operators and operands, e.g., the right child node of partial differential operators (e.g., $\partial$) must be space variables (e.g., $x$) and the left child node of $\partial$ cannot be space variables. In the specific implementation process, these constraints are applied prior to the sampling process, and the probability of the tokens is adjusted directly according to the categories of the violated constraint. In addition to imposing restrictions in the generation period, we also establish a series of regulations to double-check the generated equations and withdraw the unreasonable expressions, e.g., the coefficient of the function term is too small or the overflow of calculation. It is worth noting that regulations are aimed at removing unreasonable function terms, which is conducive to the rationality and simplicity of uncovered equations. This operation is similar to the replacement operations in genetic algorithms (i.e., the function term is set as 0), which to a certain extent increases the diversity of generated equations. Applying these constraints and regulations is convenient and extensible, and can also be incorporated with other domain knowledge for new problems.

\subsubsection*{Determine coefficients of the PDE expression.}
\paragraph{Reconstruct and split the expression tree.}
After obtaining the pre-order traversal sequence of the PDE, we first need to reconstruct it into the corresponding tree structure according to the arities of operators and operands. Then, we can split it into subtrees, i.e., the function terms, based on the plus and minus operators at the top of the expression tree, as illustrated in Fig.~\ref{fig:generation} d and e. Subsequently, we solve for the value of each function term over the entire spatiotemporal domain. In the specific solution process, $\left(u+u\right)^3$ in Fig.~\ref{fig:generation}b is taken as an instance. 
We traverse the subtree from bottom to top in a post-order traversal manner (\normalsize{\textcircled{\scriptsize{1}}}\normalsize->\normalsize{\textcircled{\scriptsize{2}}}\normalsize->\normalsize{\textcircled{\scriptsize{3}}}\normalsize), and then perform operations at each parent node (operators). At this time, the values of its corresponding child nodes have already been calculated. 

\paragraph{STRidge.}
STRidge is a widely used method in sparse regression, which can effectively determine non-trivial function terms and identify a concise equation by using the linear fit to observations. As shown in Fig.~\ref{fig:whole_framework}e, it can be utilized to solve for the coefficients of each function term based on the results in the previous step. Note that the constant $1$ is incorporated as a default constant term. 

\begin{equation}
\xi=\arg \min _{\xi}\left|\Theta(u,x) \cdot \xi-u_t\right|_2^2+\lambda|\xi|_2^2
\end{equation}
where $\lambda$ measures the importance of the regularization term. In order to prevent overfitting, we also set a threshold $tol$, and function terms with coefficients less than $tol$ will be directly ignored.
 More details can be found in PDE-FIND \cite{pde_find}. 
 
 It is worth noting that our method aims to solve the coefficients of function terms, and the possible constants in the function terms are not taken into consideration, which is also rare in PDEs. In DSR~\cite{dsr}, the constant token is introduced in the library and can be generated by agents. The constant determination is then transformed into the optimization of maximizing reward. Although it has greater flexibility (not necessarily though), the computational burden and time are increased dramatically. Our method is more efficient and more conducive to solving the PDE discovery task. For extreme special situations, we can also incorporate learnable constants in our framework.
 
\subsubsection*{Training the agent with the risk-seeking policy gradient method}

\paragraph{Reward function.} A complete representation of the generated PDEs, including function terms and their coefficients has been obtained through the above procedures. In order to effectively evaluate the generated PDE candidates, we design a reward function for the PDE discovery problem that comprehensively considers the fitness of observations and the parsimony of the equation. Assuming that the generated PDE expression is $g$, it is formulated as follows:

\begin{equation}
R=\frac{1-\zeta_1 \times n-\zeta_2 \times d }{1+RMSE},\quad RMSE=\sqrt{\frac{1}{N} \sum_{i=1}^N\left(u_{t_i}-g(u_i, x_i)\right)^2}
\end{equation}
where $n$ is the number of function terms of the governing equation; $d$ is the depth of the generated PDE expression tree; $\zeta_1$ and $\zeta_2$ are penalty factors for parsimony, which are generally set to small numbers without fine-tuning; and $N$ denotes the number of observations. It can be seen that the root mean squared error (RMSE) in the denominator evaluates the fitness of the PDE candidates to the data. The nominator is an evaluation of the parsimony. The former is designed for avoiding overfitting caused by redundant terms and the latter is primarily to prevent unnecessary structures, such as $u/u$.

\paragraph{Risk-seeking policy gradient method.} We adopt the deep reinforcement-learning training strategy to optimize the proposed agent. Herein, the generated PDE expression sequences are equivalent to the episodes in RL, and the generation process of each token corresponds to the selection of actions. Note that the total reward is based on the evaluation of the final sequence, rather than the sum of rewards for each action with a discount factor. The policy $\pi_\theta$ refers to the distribution over the PDE expression sequences $p(\tau|\theta)$. Since the generated equations with the highest reward are taken as the final result, we adopt the risk-seeking policy gradient\cite{dsr} to improve the best-case performance in the training process. The optimization objective of the risk-seeking policy gradient is the expectation of the $\left(1-\varepsilon \right)$- quantile of the rewards $\tilde{q}_{\varepsilon}(R)$, rather than the rewards of all of the samples. Its return is given by:
\begin{equation}
\setlength{\abovedisplayskip}{3pt} 
\setlength{\belowdisplayskip}{3pt}
J_{\text{risk}}(\theta ; \varepsilon) \doteq \mathbb{E}_{\tau \sim p(\tau \mid \theta)}\left[R(\tau) \mid R(\tau) \geq \tilde{q}_{\varepsilon}(R)\right]
\end{equation}
The gradient of the risk-seeking policy gradient can be estimated by:
\begin{equation}
\nabla_\theta J_{\text {risk }}(\theta ; \varepsilon) \approx \frac{\lambda_{pg}}{\varepsilon N} \sum_{i=1}^N\left[R\left(\tau^{(i)}\right)-\tilde{q}_{\varepsilon}(R)\right] \cdot \mathbf{1}_{R\left(\tau^{(i)}\right) \geq \tilde{q}_{\varepsilon}(R)} \nabla_\theta \log p\left(\tau^{(i)} \mid \theta\right)
\end{equation}
where $N$ denotes the total number of samples in a mini-batch; and $\lambda_{pg}$ measures the importance of rewards. Note that $\tilde{q}_{\varepsilon}$ is also chosen as the baseline reward  and varies by the samples at each iteration.

Additional details about the agent, hyper-parameter settings, and the optimization process can be found in Supplementary Information.

\subsection*{Acknowledgement}
This work was supported and partially funded by the National Center for Applied Mathematics Shenzhen (NCAMS), the Shenzhen Key Laboratory of Natural Gas Hydrates (Grant No. ZDSYS20200421111201738), the SUSTech – Qingdao New Energy Technology Research Institute, and the National Natural Science Foundation of China (Grant No. 62106116).

\subsection*{Code availability}
The implementation details of the whole process are available on GitHub at https://github.com/menggedu\\/DISCOVER.

\subsection*{Data availability}
All of the data used in the experiments are available on GitHub at https://github.com/menggedu/DISCOVER.


\bibliography{bibliography}
\bibliographystyle{naturemag}

\end{document}


\maketitle

\beginsupplement

\captionsetup[figure]{labelformat=simple, labelsep=period}
\captionsetup[table]{labelformat=simple, labelsep=period}
The supplementary document provides a detailed description of the proposed framework in terms of both method and experimental details. In the method, we mainly supplement the following five aspects, including (1) the introduction of algorithms in DISCOVER; (2) the mechanism of the monotonic attention; (3) details of customized constraints and regulations; (4) the risk-seeking gradient method: and (5) hyperparameter settings. In the experiment parts, we first provide the background and details of the used data set and then discuss the optimization process and effect of noise levels and data volume. The structure-aware agent designed in DISCOVER is further compared with the standard LSTM agent and shows great superiority in capturing local information and long-distance information. Finally, we discuss the effect of the number and utilization of generated samples on the efficiency of mining equations.
\section{Methods}

\subsection{Algorithm.}
Herein, we provide the algorithm details of DISCOVER. In the implementation process, a priority queue $Q$ is introduced to store the fixed number of $K$ PDE expressions with the best reward. It can be dynamically updated at each iteration. Whenever there are rewards higher than the internal expressions, the new expressions enter the queue and expressions with lower rewards are omitted. Users can choose the best expressions by balancing accuracy and parsimony.
\begin{algorithm}[!ht]
\caption{Identifying the open-form PDE from data with DISCOVER}
\KwIn{symbol library: $\mathcal{L}$; time derivatives: $\mathbf{U}_t$; total generated expressions at each iteration: $N$; reward threshold: $\mathcal{R}^*$; learning rate: $\alpha$; number of retained expressions in priority queue: $K$.
}
\KwOut{The best PDE expression $\tau^{*}$.}
\textbf{Initialize:} The structure-aware LSTM agent with parameters $\theta$; priority queue $Q$. 

\textbf{repeat}\\
1. Generate $N$ PDE expression traversals $\mathcal{T}=\left\{\tau_i \right\}^{i=1}_N$.


2. Restore $\mathcal{T}$ into the tree structure and split PDE traversals into function terms $\Theta$. 

3. Calculate the coefficients of the function terms $\hat{\xi}$.
\begin{center}
$\hat{\xi}=\operatorname{argmin}_{\xi}\left\|\Theta \xi-\mathbf{U}_{t}\right\|_{2}^{2}+\lambda\|\xi\|_{2}^{2}$
\end{center}
\vspace{-0.5em}
4. $\mathcal{T}_{val} \gets \mathcal{T}$.  \Comment{Filtering the PDE expression traversals with predefined regulations.}

5. Compute rewards $\mathcal{R}$ and the $1-\varepsilon$ quantile of rewards $\mathcal{R}_{\varepsilon}$.

6. $\mathcal{T}_{\varepsilon} \gets \left\{\tau_{i} \in \mathcal{T}_{val}: \mathcal{R}(\tau_i) \geq \mathcal{R}_{\varepsilon} \right\}$. \Comment{Select the traversals whose rewards larger than $\mathcal{R}_{\varepsilon}$.}

7. $\theta \gets \theta + \alpha(\nabla_\theta J_{\text {risk }}+\nabla_\theta J_{\text {entropy }})$. \Comment{Update the agent with gradient ascent method.}

8. Update $Q$ and keep $K$ PDE expressions with maximum rewards.

\textbf{until} The maximum reward of $Q$ is larger than $\mathcal{R}^*$.
\end{algorithm}

As shown in Algorithm~\ref{Rebuild}, when PDE expression traversals are generated by the structure-aware agent, it is necessary to first reconstruct it into a tree and return the root node of it. After that, we parse the tree further and decompose it into subtrees according to the predefined addition and subtraction operators, as shown in Algorithm~\ref{splittree}. Based on these two algorithms, we can obtain the representation of all of the function terms and then remove the function terms and PDE candidates that violate the regulations. Finally, coefficients of legal terms can be identified by STRidge~\citesupp{pde_find}.
\begin{algorithm}[!ht]
\caption{Rebuilding the tree structure from the pre-order traversal}
\label{Rebuild}
\KwIn{ $\tau$ \Comment{A PDE expression traversal.}}
\KwOut{ Root  \Comment{ Root node of the generated tree.} }
\SetKwFunction{FMain}{Rebuild\_tree}
    \SetKwProg{Fn}{Function}{:}{}
    \Fn{\FMain{$\tau$}}{
$S=[]$ \Comment{Store non-leaf nodes}\;
full\_node = None \Comment{The node whose arity is currently 0}\;
\While{$len(N) \neq 0$ or $len(S) \neq 0$}{

  \eIf{full\_node is None}{
    node = $\tau$.pop(0)\;
    arity=node.arity
  }{node=full\_node\;
  arity=0
  }
  \eIf{arity$\geq 0$}{
    $S \gets (node,arity)$\;
  }{
      full\_node=node\;
      \If{$len(S) \neq 0$}{
        last\_op, last\_arity = $S.pop(-1)$\;
        last\_op.children$\gets node$\;
        last\_arity-=1\;
        \eIf{last\_arity$\geq 0$}{
            $S \gets (last\_op,last\_arity)$ \;
            full\_node=None
        }{
        full\_node=last\_op
        }
      }
  }
}
    \Return full\_node
}
\end{algorithm}


\begin{algorithm}[!ht]
\caption{Splitting the PDE expression tree into function terms}
\label{splittree}
\KwIn{root.\Comment{Root node of the PDE expression tree.}}
\KwOut{ List of root node of subtrees.}
\SetKwFunction{FMain}{SplitTree}
    \SetKwProg{Fn}{Function}{:}{}
    \Fn{\FMain{root}}{

value = root.val \Comment{ Fetch the symbol of root node.}\;
\If{ value not in ['$+$','$-$']}{
\Return [root] \Comment{Terminate the split operation if the current node does not have a plus or minus symbol.}}

\Return [SplitTree(root.children[0]), SplitTree(root.children[1])]
}
\end{algorithm}

\subsection{Monotonic attention.}
\label{sup_ma}

\begin{figure}[htbp]
\centering
\includegraphics[width=0.8\linewidth]{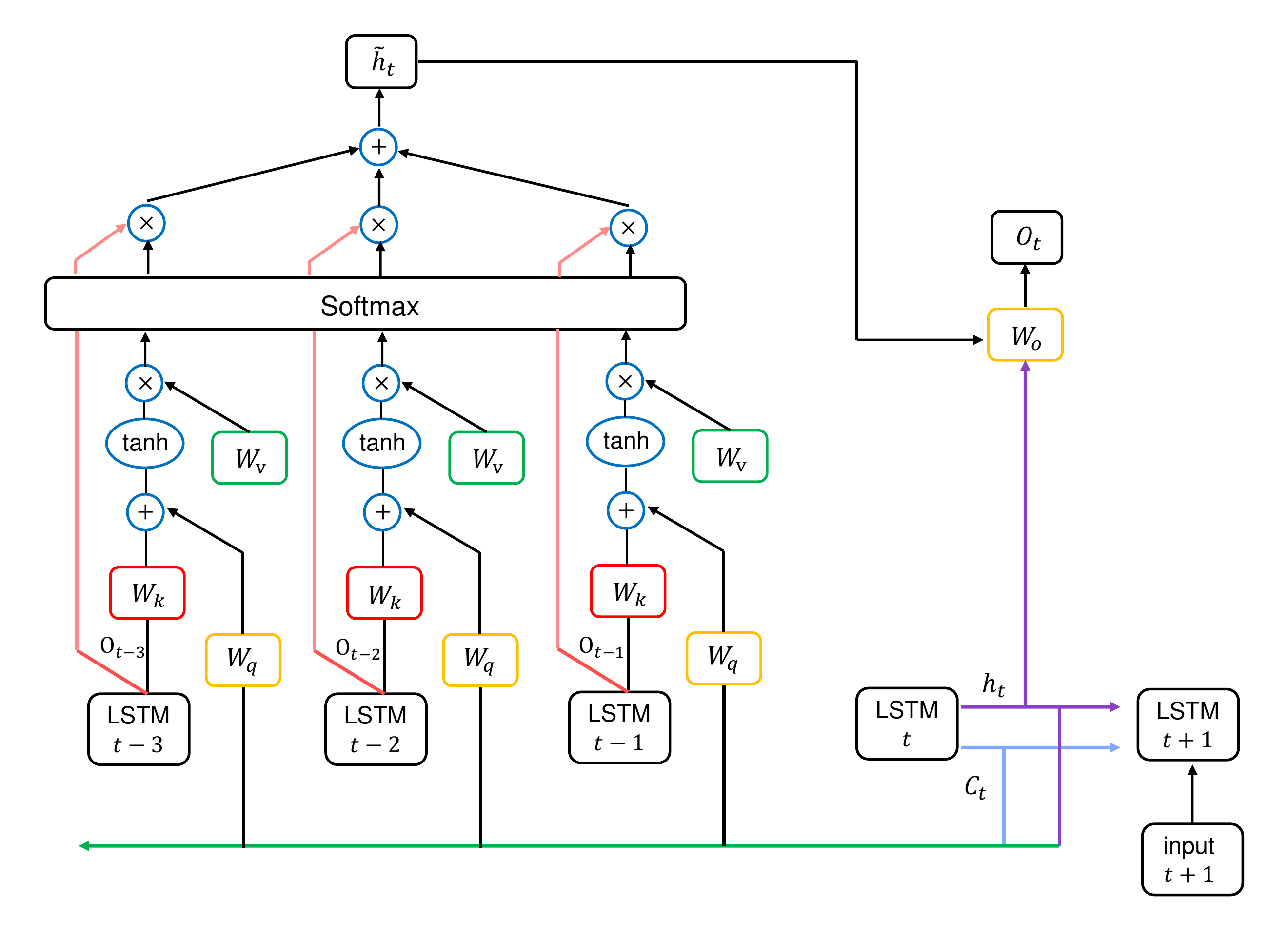}
\caption{\textbf{The architecture of monotonic attention.}
Only the information of the past three time steps is integrated here. In fact, the time-span can be set longer, so that it can focus on further historical information.}
\label{fig_ma}
\vspace{-1.0em}
\end{figure}

It is well known that the LSTM recurrent neural network is effective at dealing with sequence problems, such as machine translation~\citesupp{lstm_mt} and text generation~\citesupp{lstm_tg}. The history information is recursive-compressed  and stored in a memory cell, and three gates, including an input gate, an output gate, and a forget gate, control the flow of the information and decision-making. However, its constraint is obvious, i.e., prediction of the current time-step largely depends on the adjacent units, and recursively updating the information destroys the structural information of the input. Furthermore, the storage capacity of memory cells is limited, and information loss increases as the sequence grows. Some attempts have been made to incorporate soft attention to comprehend internal structures and directly select useful information from the previous tokens \citesupp{lstm_att1,lstm_att2,attention_wrapper}. 

In order to capture the long-distance dependencies and structural information, we wrapped the original LSTM with a monotonic attention layer (MAL), which mainly draws on the architecture and configuration in LSTMN \citesupp{attention_wrapper}. Attention memory is utilized here to simulate the human brain to read historical information and mine the relationship between them. The architecture of MAL is illustrated in Fig.~\ref{fig_ma}. Compared with the standard LSTM, an extra hidden vector $\tilde{h_t}$ is introduced to store the relations between tokens. At time-step $t$, an attention function can be calculated by: 

\begin{equation}
\left[h_{t}, c_{t}\right]=\operatorname{LSTM}\left(x_{t},\left[h_{t-1}, c_{t-1}\right]\right)
\end{equation}
\begin{equation}
a_{i}^{t}=W_{v}^{T} \tanh \left(W_{q}\left[h_{t}, c_{t}\right]+W_{k}\mathrm{O}_{i}\right)
\end{equation}
\begin{equation}
s_{i}^{t}=\operatorname{softmax}\left(a_{i}^{t}\right)
\end{equation}

where $W_q$, $W_k$, and $W_v$ are the parameter matrices used to linearly project queries, keys, and their output, respectively; and the state vector $s_t$ denotes a probability distribution over the previous inputs to measure the degree of attention to historical information. It is also used to calculate our new hidden vector. By combining $\tilde{h_t}$ containing the relation information and the original hidden vector $h_t$, the final output can be represented as follows:
\begin{equation}
\tilde{h}_{t}=\sum_{i}^{t-1} s_{i}^{t} \mathrm{O}_{i} 
\end{equation}
\begin{equation}
\mathrm{O}_{t}=W_{o}\left[h_{t}, \tilde{h}_{t}\right]
\end{equation}

The specific implementation refers to the source code in TensorFlow \citesupp{tensorflow}, and some modifications are made to better combine structured input.

\subsection{Constraints and regulations.}
The main purpose of constraints and regulations is to prevent the framework from generating equations that violate the laws of mathematics and physical laws, and reduce the search space. Among them, constraints are directly applied to the probability distribution of the symbol library at the regressive generation process. The probability of an unreasonable symbol is reduced to a small number, even 0. Constraints in the framework include:
\begin{enumerate}[1)]
\item Total length of the sequence is less than 64.
\item Number of plus and minus operators is less than 10. 
\item Relationship limits between differential operators and spatial inputs: the right child node of partial differential operators (e.g., $\partial$) must be space variables (e.g., $x$); the left child node of $\partial$ cannot be space variables, and the plus and the minus operator cannot appear in the descendants of $\partial$ (optional).
\end{enumerate}
Regulations are applied to the function terms of generated PDE expression, including:
\begin{enumerate}[1)]
\item Descendants of partial differential operators do not contain state variables (e.g., $u$).
\item Arithmetic underflow and overflow.
\item Numerical errors during conducting sparsity-promoting methods.
\item Coefficient of the corresponding function term is less than a default number (e.g., $1\times10^{-5}$).
\end{enumerate}

\subsection{Risk-seeking policy gradient method.}
Different from PDE-Net 2.0 \citesupp{pde_net2}, in this problem, we cannot directly establish the computational graph between rewards and generated expression trees, and utilize gradient descent to update the model parameterized by $\theta$. Therefore, we adopt the deep reinforcement-learning training strategy. Specifically, the generated PDE expression sequences are equivalent to the episodes in RL, and the generation process of each token corresponds to the selection of actions. It is worth noting that the total reward is not the sum of each action with a discount factor, but rather is based on the evaluation of the final sequence. The policy $\pi_\theta$ refers to the distribution over the PDE expression sequences $p(\tau|\theta)$. The standard policy gradient is a risk-neural method, i.e., its goal is to maximize the expectation of the generated policy, and the return is as follows:
\begin{equation}
J\left(\pi_\theta\right)=\mathbb{E}_{\tau \sim P^{\pi_\theta}}[R(\tau)]
\end{equation}
However, for certain problems, such as PDE discovery, our intention is to ensure that the best-case sample is adequate for the problems,  which is similar to symbolic regression \citesupp{sr1,sr2} and neural architecture search (NAS) \citesupp{nas}. By referring to the method of risk-seeking policy gradient in DSR \citesupp{dsr}, we train agents to improve the best-case performance, instead of the average performance, to alleviate the mismatch between the objective and performance evaluation. The idea of this method comes from a well-known risk measure, conditional value at risk (CVaR), defined as 
$CVaR_\varepsilon(R)=\mathbb{E}\left[R \mid R \leq q_\varepsilon(R)\right]$, where $q_\varepsilon$   is the $\varepsilon$ quantile of the rewards. It is designed to improve the worst samples in the current policy and avoid risks, and is usually applied in vehicle driving or finance\citesupp{cvar0,cvar1,cvar2}. In contrast, the optimization objective of risk-seeking is the expectation of the $\left(1-\varepsilon \right)$- quantile of the rewards $\tilde{q}_{\varepsilon}(R)$. Its return is given by:
\begin{equation}
\setlength{\abovedisplayskip}{3pt} 
\setlength{\belowdisplayskip}{3pt}
J_{\text{risk}}(\theta ; \varepsilon) \doteq \mathbb{E}_{\tau \sim p(\tau \mid \theta)}\left[R(\tau) \mid R(\tau) \geq \tilde{q}_{\varepsilon}(R)\right]
\end{equation}
The gradient of the risk-seeking policy gradient can be estimated by:
\begin{equation}
\nabla_\theta J_{\text {risk }}(\theta ; \varepsilon) \approx \frac{\lambda_{pg}}{\varepsilon N} \sum_{i=1}^N\left[R\left(\tau^{(i)}\right)-\tilde{q}_{\varepsilon}(R)\right] \cdot \mathbf{1}_{R\left(\tau^{(i)}\right) \geq \tilde{q}_{\varepsilon}(R)} \nabla_\theta \log p\left(\tau^{(i)} \mid \theta\right)
\end{equation}
where $N$ denotes the total number of samples in a mini-batch; and $\lambda_{pg}$ measures the importance of rewards. Note that $\tilde{q}_{\varepsilon}$ is also chosen as the baseline reward  and varies by the samples at each iteration.
In addition, based on maximum entropy reinforcement-learning \citesupp{soft_entro}, the entropy value of each output action under the current policy is also required to be maximized to prevent generating a certain action continuously. The gradient of entropy can be expressed by:
\begin{equation}
\nabla_\theta J_{\text {entropy }}(\theta ; \varepsilon) \approx \frac{1}{\varepsilon N} \sum_{i=1}^N\left(\lambda_{\mathcal{H}} \mathcal{H}(\tau^{(i)}\mid \theta)\right)
\end{equation}
where $\lambda_{\mathcal{H}}$ is the temperature parameter that controls the relative importance of the entropy term against the reward. By combining these two parts, the agent can be optimized to iteratively improve the best-case performance, while increasing the exploration ability to avoid getting stuck in local optima.

\subsection{Hyperparameters.}
\label{append_hyper}

The default hyperparameters used to mine the above PDEs are shown in Table~\ref{hyperp}. As mentioned above, we use the plus or minus operator that appears at the top level of the tree as the identifier to split the equation into several function terms. We require that the number of their occurrences should not exceed five, which means that the generated expressions can only be spliced into at most six function terms. The parsimony penalty factor which guarantees the simplicity of the equation is set to 0.01. In each iteration, the agent will generate a total of $N=500$ PDE expressions. Finally, after filtering the illegal expressions and low rewards, only 2\% ($\varepsilon=0.02$) of the total expressions, i.e., 10 equations with the highest reward, are selected for the update of the agent. In the optimization process, the coefficients of entropy loss and policy gradient loss are set to 0.03 and 1, respectively.

\begin{table}[htbp]
  \caption{\textbf{Default hyperparameter settings for discovering open-form PDEs.}}
  \label{hyperp}
  \centering
\setlength{\abovecaptionskip}{0cm} 
\setlength{\belowcaptionskip}{0.2cm}
\begin{tabular}{ccc}
\hline
\textbf{Hyperparameter} &  \textbf{Default value} &  \textbf{Definition}                            \\ \hline
$N_{subtree}$     & 6             & Maximum number of function terms               \\ 
$D_{subtree}$    & 8 & Maximum depth of subtrees
\\
$\zeta_1$           & 0.01          & Parsimony penalty factor for redundant function terms             \\
$\zeta_2$           & 0.0001          & Parsimony penalty factor for unnecessary structures             \\
$N$        & 500           & Total generated expressions at each iteration         \\
$\varepsilon$        & 0.02          & Threshold of reserved expressions     \\
$\lambda$           & 0             & Weight of the STRidge regularization term    \\
$tol$           & $1\times10^{-4}$       & Threshold of weights for reserved function terms  \\
$\lambda_{\mathcal{H}}$         & 0.03          & Coefficients of entropy loss          \\
$\lambda_{pg}$          & 1             & Coefficients of policy gradient loss  \\ 
$T$ & 20 & Time span of the monotonic attention \\\hline
\end{tabular}
 
\end{table}

\section{Experiments}

\subsection{Data description.}
The data used in this paper are mainly divided into two categories: 1D and 2D systems. The data descriptions of the 1D canonical systems can be found in SGA~\citesupp{sga_pde}. To verify the effectiveness of our framework in mining high-dimensional and high-order nonlinear systems, we introduce the Allen-Cahn and Cahn-Hilliard equations in 2D dynamic systems. They are originally introduced to describe the non-conservative and conservative phase variables in the phase separation process, respectively. Both models are recognized as gradient flow systems. Assume that the Ginzburg–Landau free energy functional takes the following form:
\begin{equation}
F=\int_\omega \frac{\gamma_1}{2}|\nabla u|^2+\frac{\gamma_2}{4}\left(u^2-1\right) d \mathbf{x}
\end{equation}

The Allen-Cahn equation can be obtained as an $L^2$ gradient flow with the following form:
\begin{equation}
u_t =\gamma_{1}\Delta u+\gamma_{2}(u-u^3)
\end{equation}
where $\gamma_{1}=0.001$ and $\gamma_{2}=1$.
The Cahn-Hilliard equation can be obtained as a $H^{-1}$ gradient flow as shown below:
\begin{equation}
u_t =\Delta(-\gamma_{1}\Delta u+\gamma_{2}(u-u^3))
\end{equation}
where $\gamma_{1}=0.5$ and $\gamma_{2}=-1$.
The specific discretization details of the two equations are shown below:

(1) Allen-Cahn equation: 
The spatial domain is taken as $\textbf{x}\in[0,1]^2$ with 64 points along each axis. The temporal domain is taken as $t\in(0,5]$ with 100 points for discretization in timescale. The initial condition is set as $u(0,x_1,x_2) = sin(4\pi x_1)cos(4\pi x_2)$, where $(x_1,x_2)$ are discrete points in the spatial domain. The periodic boundary conditions are imposed with $u^{d}(t,-1) = u^{d}(t,1)$, for $d=0,1$.

(2)	Cahn-Hilliard equation:
The spatial domain is taken as $\textbf{x}\in[0, 64]^2$ with 64 points along each axis. The temporal domain is taken as $t\in(0, 10]$ with 500 points for discretization in the timescale. The random noisy initial condition is set as $u(0,x_1,x_2)=rand()-0.5$,where $(x_1,x_2)$ are discrete points in the spatial domain. The same periodic boundary conditions are imposed as those in the Allen-Cahn equation.

\begin{figure}[!ht]
\centering
\includegraphics[width=0.8\linewidth]{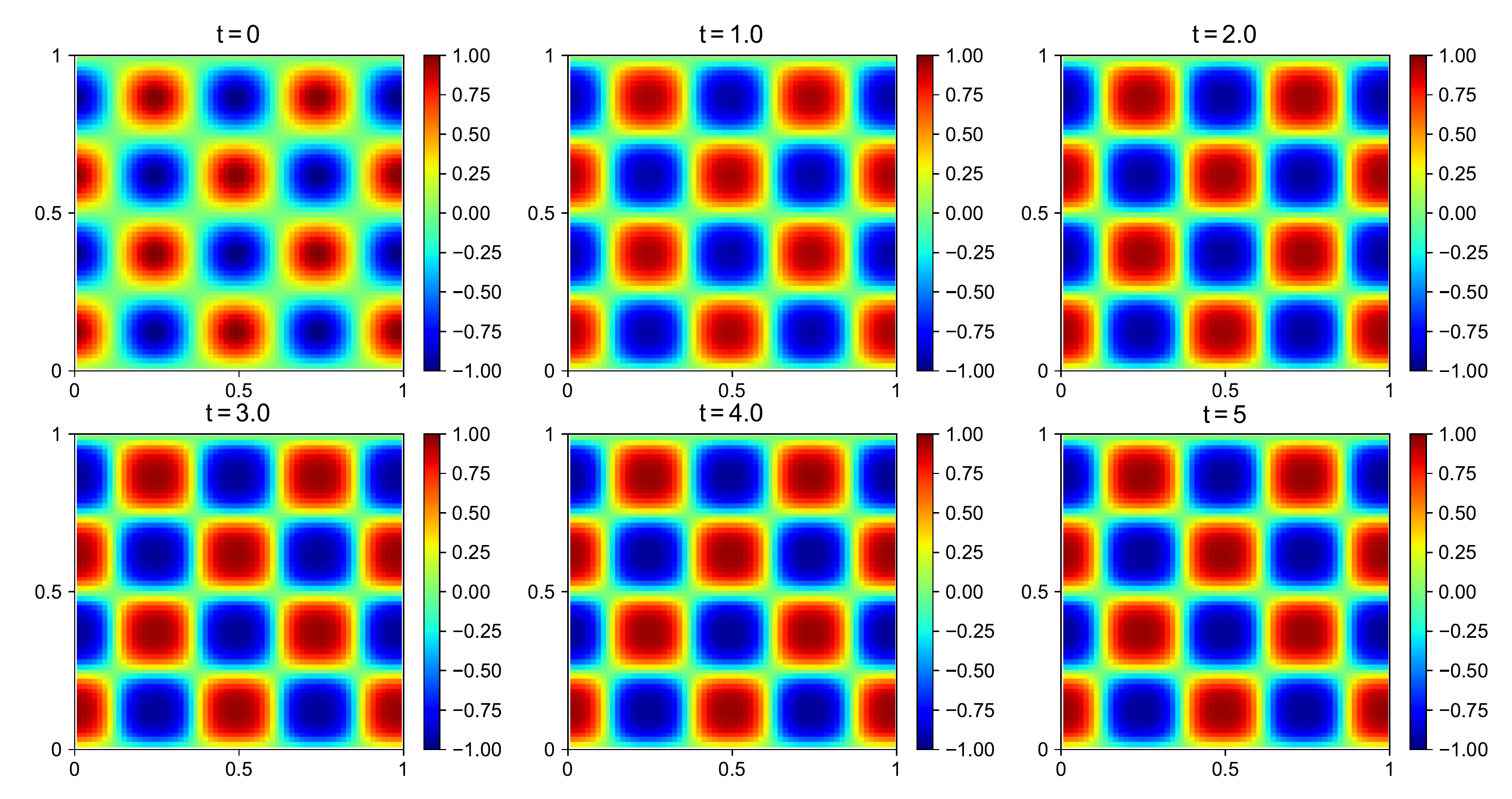}
\caption{\textbf{Temporal morphology for the Allen-Cahn equation.}}
\label{fig:ac_data}
\end{figure}

\begin{figure}[!ht]
\centering
\includegraphics[width=0.8\linewidth]{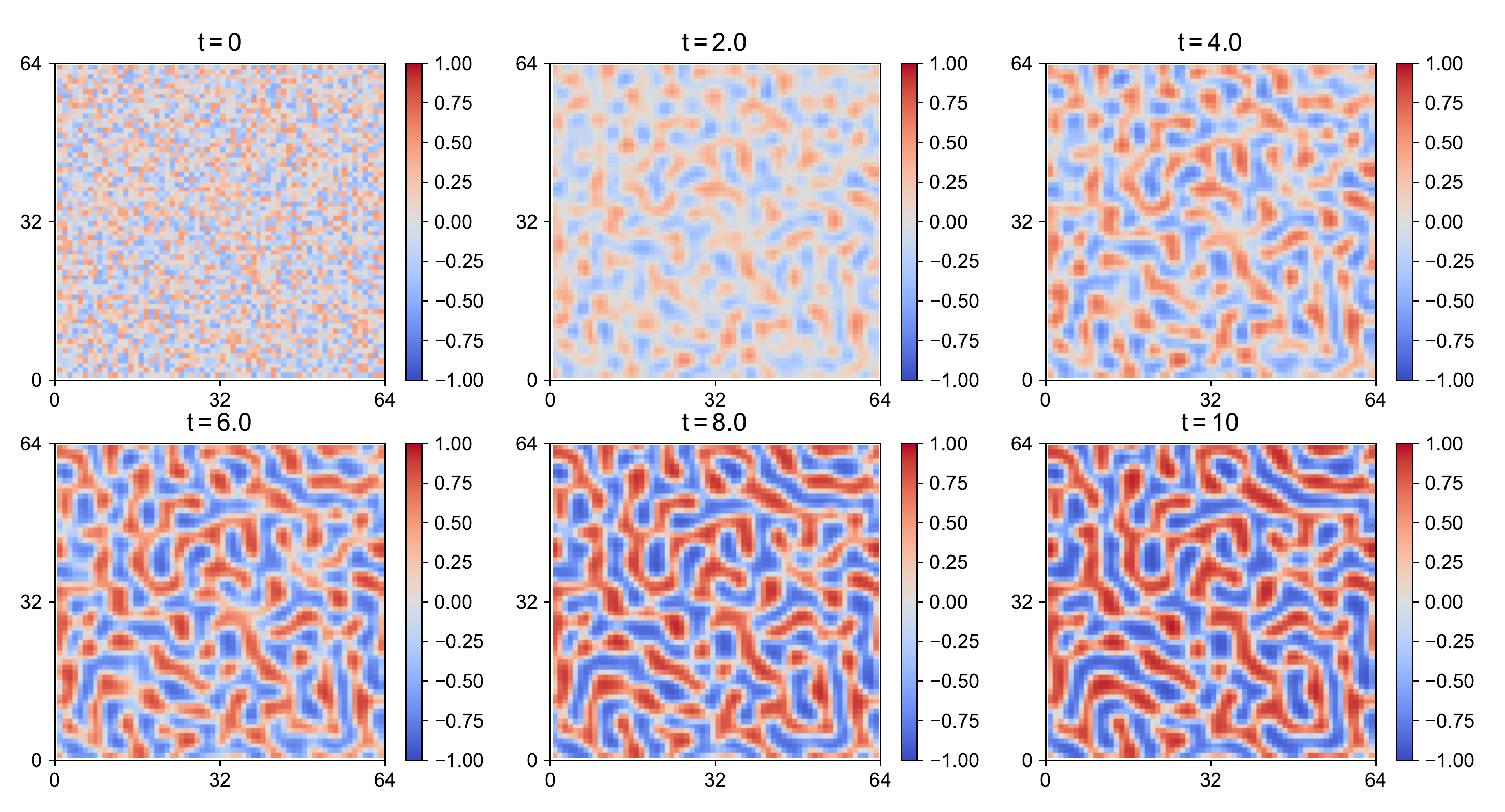}
\caption{\textbf{Temporal morphology for the Cahn-Hilliard equation.}}
\label{fig:ch_data}
\end{figure}

\subsection{The optimization process of DISCOVER.}

\begin{figure}[!ht]
\centering
\includegraphics[width=\linewidth]{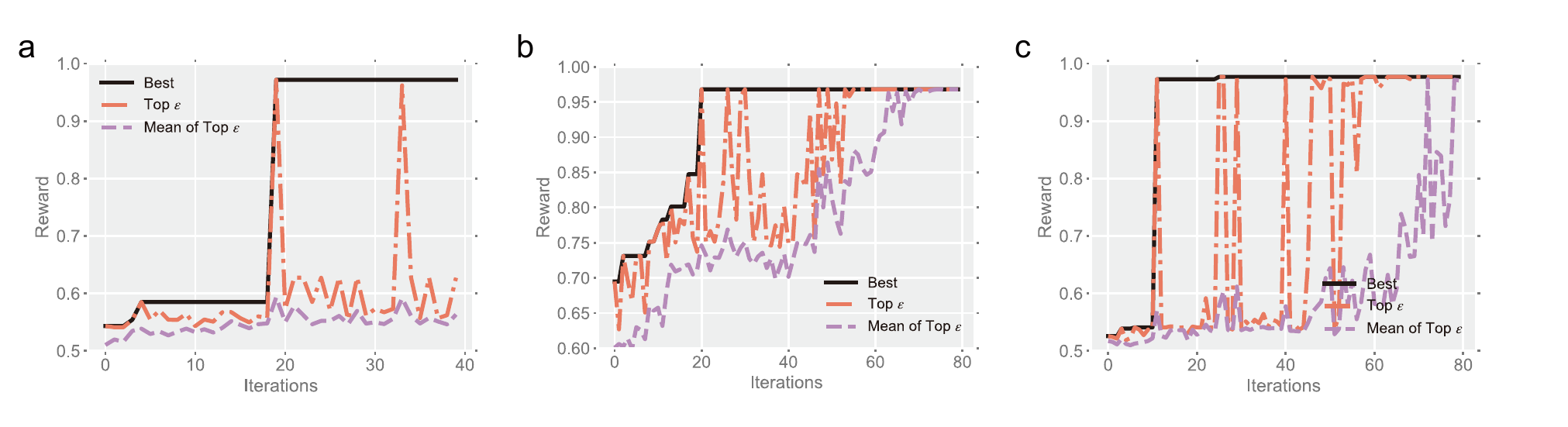}
\caption{\textbf{Rewards distribution of the best case of all time, top $\epsilon$, and mean of top $\epsilon$ of the samples.} \textbf{a} Burgers' equation. \textbf{b} The Chafee-Infante equation. \textbf{c} The KdV equation.}
\label{fig:result1d}
\end{figure}

\begin{figure}[!ht]
\centering
\includegraphics[width=\linewidth]{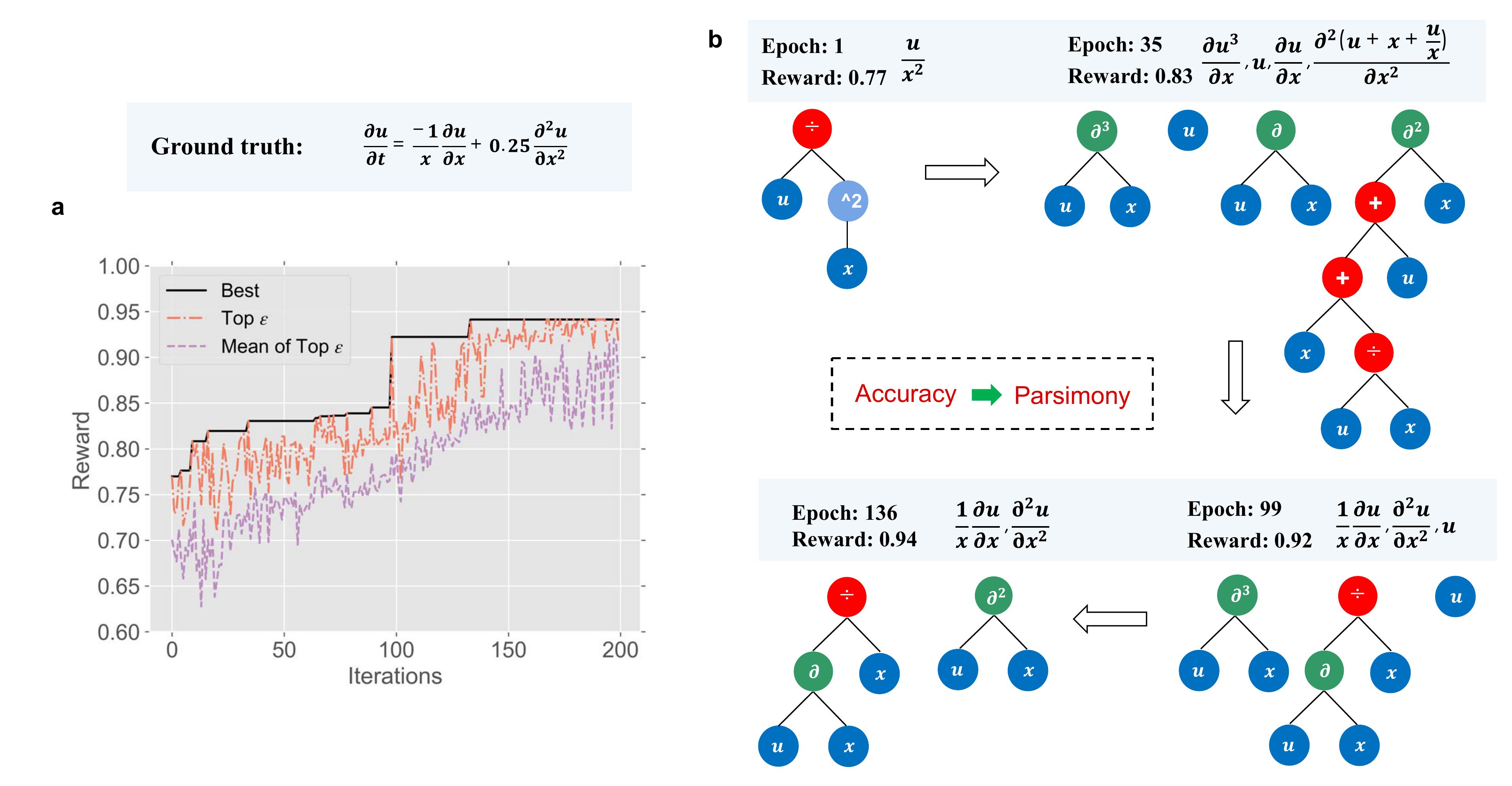}
\caption{\textbf{The optimization process of PDE\_divide.} \textbf{a} Reward distribution. \textbf{b} Evolution of the function terms.}
\label{fig:opt_process}
\end{figure}
\begin{figure}[!ht]
\centering
\includegraphics[width=\linewidth]{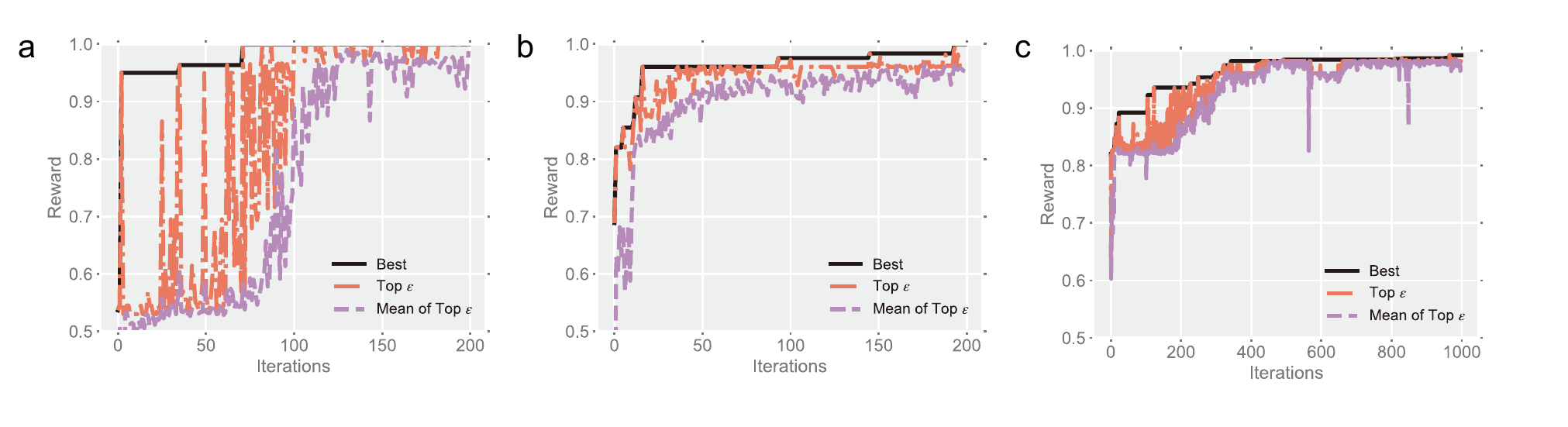}
\caption{\textbf{Rewards distribution of the best case of all time, top $\epsilon$, and mean of top $\epsilon$ of the samples.} \textbf{a} The Allen-Cahn equation. \textbf{b} The Cahn-Hilliard equation (with Laplace operator). \textbf{c} The Cahn-Hilliard equation (without Laplace operator). }
\label{fig:result2d}
\end{figure}

To demonstrate the concrete details of the optimization process, we provide the reward distribution of the best-performing, top $\varepsilon$, and the mean of the top $\varepsilon$ in each batch during the training process for different systems. It can be seen from Fig.~\ref{fig:result1d} and Fig.~\ref{fig:opt_process} that as the iteration progresses, rewards of the best-case increase in a stepwise manner, and there is a sudden spike for the Burgers' equation and the Kdv equation. The risk-seeking policy gradient method is aimed at improving the better-case performance, and thus it is not necessary for all of the generated equation forms to have high rewards to uncover the true form. Note that the PDE\_compound equation is relatively simple, and the correct equation form can be identified in the first few rounds of iterations. The PDE\_divide is taken as an example to show the evolution process of the generated equation expressions during the iteration process as shown in Fig.~\ref{fig:opt_process}b. It can be seen that from the 1st to the 99th iteration, the composition of the equation terms gradually approaches the correct one. The increase in reward gradually benefits from the increase in accuracy. From the $99^{th}$ iteration to the $136^{th}$ iteration, in addition to a further gain of accuracy, the parsimony of the expressions is also considered. Therefore, in the process of revealing equations from the data, when the accuracy of the generated terms is similar, the expressions with fewer terms have a bigger reward. This also ensures that the final equation form is both accurate and parsimonious.
Fig.~\ref{fig:result2d} illustrates the reward distribution of the 2D systems. For the Cahn-Hilliard equation, 
the library that newly defines the Laplace operator allows the agent to identify the correct equation in 200 iterations, which can be seen in Fig.~\ref{fig:result2d}b. When only the first- and second-order differential operators are used, high-dimensional operations can only be accomplished by the nesting of differential operations. The numerically equivalent equation form is verbose and it takes more than four times the number of iterations of the former one. The identified equation form is shown below:
\begin{equation}
   u_t = 0.999(u^3)_{xx}+0.999(u^3)_{yy}-0.999u_{xx}-1.000u_{yy}-0.499(u_{xx})_{xx}-0.500(u_{xx})_{yy}
\end{equation}

\subsection{Effect of noise levels and measurement points.}

Since observations obtained from real scenes are often sparse and noisy, it is necessary to test whether DISCOVER can mine the correct form of equations under different volumes of data and noise levels. During the evaluation stage, the finite difference method is utilized to calculate derivatives, which are sensitive to noise~\citesupp{DL_PDE}. Therefore, we built a fully connected neural network as a surrogate model to smooth available noisy data, and at the same time, metadata at other spatial-temporal locations could be generated. It is worth noting that PDE-FIND \citesupp{pde_find} utilizes polynomial interpolation to deal with noisy data, but this method is only applicable to a complete library. All of the derivatives can be obtained by a one-time calculation. However, the calculation burden is unbearable for our framework since variable representations require manifold derivative calculations. Moreover, it cannot handle the boundary points well. The metadata method that we use is relatively simple and time-saving, and only involves the pre-processing part. After smoothing the noisy data, automatic differentiation can be incorporated to calculate the derivatives.

In order to preprocess the measurements, a fully connected feedforward neural network (FNN) is constructed to map the system inputs (i.e., x and t) to its output $u$. For different systems, we adopt the same network structure, i.e., there are three hidden layers between the input layer and the output layer, and the number of neurons in each layer is 64. Sin(x) is taken as the activation function. The mean square error (MSE) loss function was developed to evaluate the difference between the measurement and the model output. The number of epochs is set to 100,000. In addition, during training, the training set and validation set are divided according to the ratio of 8:2, and the early stopping strategy is adopted to avoid over-fitting. A series of smooth predictions generated by FNN will be used to solve derivatives.

The relative $l_2$ error defined as $|| \hat\xi - \xi_{true}||_1 /||\xi_{true}||_1$ is utilized to assess the accuracy of identified coefficients of function terms. 
The identified equations and error rates for different systems under different noise levels are shown in Supplementary Table~\ref{table:noise_res}. Only 80\% of the total measurements were randomly sampled and utilized to train the surrogate model. It can be seen that DISCOVER is able to identify the correct equation form of the KdV and Burgers' equations with a relatively small error when the data are added to the maximum 5\% noise. For the other three equations, satisfactory results can be obtained only under the 1\% noise condition. It is worth noting that for 2D cases, although the correct equation structure of the Allen-Cahn equation can be identified with 1\% noise, the error of the determined coefficients increases significantly. For the Cahn-Hilliard equation, due to the existence of the fourth-order derivative and compound structure, the correct equation form cannot be found with current settings when only noisy data are available. 

It is worth noting that our method is far from perfect. Due to the sensitivity of numerical differentiation to noise, it is still unable to handle high-noise cases despite a certain degree of smoothing. Although automatic differentiation can alleviate this problem to a large extent, combining automatic machine learning methods and symbolic representation is required to ensure the derivation of any form of function term. At the same time, in the future, physical constraints need to be incorporated in FNN based on physical knowledge of the uncovered equations, so as to enhance the robustness of the model to noise and reduce the dependence on data volume.
\begin{table}[!ht]
\setlength{\abovecaptionskip}{0cm} 
\setlength{\belowcaptionskip}{0.2cm}
\renewcommand\arraystretch{1.5}
\caption{\textbf{Summary of discovered results for different PDEs of mathematical physics under different noise levels.}}
\label{table:noise_res}
\centering
\begin{tabular}{cccc}

\toprule
\multirow{5}{*}{KdV}            & \multicolumn{3}{c}{Correct PDE:\quad $u_t=-u\times u_x-0.0025u_{xxx}$}                                                                     \\ \cline{2-4} 
                                & Noise level                     & Identified PDE                     & Error(\%)                     \\ \cline{2-4} 
                                & Clean data                      &    $u_t=-0.5001(u\times u)_x-0.0025u_{xxx}$                                &$0.09\pm0.07$                               \\
                                & 1\% noise                       &     $u_t=-0.4983(u\times u)_x-0.0025u_{xxx}$                               &  $0.31\pm0.04$                             \\
                                & 10\% noise                       &  $u_t=-0.9748u\times u_x-0.0024u_{xxx}$                                  &   $2.50\pm0.028$  
                                \\\hline
\multirow{6}{*}{Burgers}        & \multicolumn{3}{c}{Correct PDE:\quad$u_t=-uu_x+0.1u_{xx}$  }                                                                     \\ \cline{2-4} 
                                & Noise level                     & Identified PDE                     & Error(\%)                     \\\cline{2-4} 
                                & Clean data                      &  $u_t=-1.0010uu_x+0.1024u_{xx}$                                   &   $1.25\pm1.61$                            \\
                                & 1\% noise                       &   $u_t=-0.4992(u\times u)_x+0.0982u_{xx}$                                  &  $0.95\pm1.12$                           \\
                                & 10\% noise                       &   $u_t=-0.4886(u\times u)_x+0.0943u_{xx}$   &       $3.94\pm2.35$                        \\ 
                                \hline
\multirow{4}{*}{Chafee-Infante} & \multicolumn{3}{c}{Correct PDE:\quad$u_t=u_{xx}+u-u^3$ }                                                                     \\ \cline{2-4} 
                                & Noise level                     & Identified PDE                     & Error(\%)                     \\\cline{2-4}
                                & \multicolumn{1}{c}{Clean data}  & \multicolumn{1}{c}{$u_t=1.0002u_{xx}-1.0008u+1.0004u^3$}               & \multicolumn{1}{c}{$0.04\pm0.03$}          \\
                                & \multicolumn{1}{c}{1\% noise}   & \multicolumn{1}{c}{$u_t=0.9866u_{xx}-0.9862u+0.9877u^3$}               & \multicolumn{1}{c}{$1.32\pm0.08$}          \\ 
                                & \multicolumn{1}{c}{10\% noise}   & \multicolumn{1}{c}{$u_t=0.9777u_{xx}-0.9766u+0.9869u^3$}               & \multicolumn{1}{c}{$1.96\pm0.56$}          \\                                 \hline
\multirow{4}{*}{PDE\_compound}  & \multicolumn{3}{c}{Correct PDE:$u_t=(uu_x)_x$ }                                                                     \\ \cline{2-4} 
                                & \multicolumn{1}{c}{Noise level} & \multicolumn{1}{c}{Identified PDE} & \multicolumn{1}{c}{Error(\%)} \\\cline{2-4} 
                                & \multicolumn{1}{c}{Clean data}  & \multicolumn{1}{c}{$u_t=0.5002(u^2)_{xx}$}               & \multicolumn{1}{c}{$0.04$}          \\
                                & \multicolumn{1}{c}{1\% noise}   & \multicolumn{1}{c}{$u_t=0.5003(u^2)_{xx}$}               & \multicolumn{1}{c}{$0.05$}          \\ \hline
\multirow{4}{*}{PDE\_divide}    & \multicolumn{3}{c}{Correct PDE:\quad$u_t=-u_x/x+0.25u_{xx}$  }                                                                     \\ \cline{2-4} 
                                & \multicolumn{1}{l}{Noise level} & \multicolumn{1}{c}{Identified PDE} & \multicolumn{1}{c}{Error(\%)} \\\cline{2-4} 
                                & \multicolumn{1}{c}{Clean data}  & \multicolumn{1}{c}{$u_t=-0.9979u_x/x+0.2498u_{xx}$ }               & \multicolumn{1}{c}{$0.14\pm0.10$}          \\
                                & \multicolumn{1}{c}{1\% noise}   & \multicolumn{1}{c}{$u_t=-0.9803u_x/x+0.2478u_{xx}$}               & \multicolumn{1}{c}{$1.42\pm0.78$}          \\ \hline
\multirow{4}{*}{Allen-Cahn}    & \multicolumn{3}{c}{Correct PDE:\quad $u_t=0.001\Delta u-u^3+u$ }                                                                     \\ \cline{2-4} 
                                & \multicolumn{1}{l}{Noise level} & \multicolumn{1}{c}{Identified PDE} & \multicolumn{1}{c}{Error(\%)} \\\cline{2-4} 
                                & \multicolumn{1}{c}{Clean data}  & \multicolumn{1}{c}{$u_t=0.001(u_{xx}+u_{yy})-0.999u^3+1.000u$ }               & \multicolumn{1}{c}{$0.025\pm0.05$}          \\
                                & \multicolumn{1}{c}{1\% noise}   & \multicolumn{1}{c}{$u_t=0.0007(u_{xx}+u_{yy})-0.884u^3+0.870u$}               & \multicolumn{1}{c}{$21.72\pm10.94$}          \\ \hline
\end{tabular}
\end{table}

\subsection{Comparison to DSR.}
\begin{figure}[htbp]
\includegraphics[width=\linewidth]{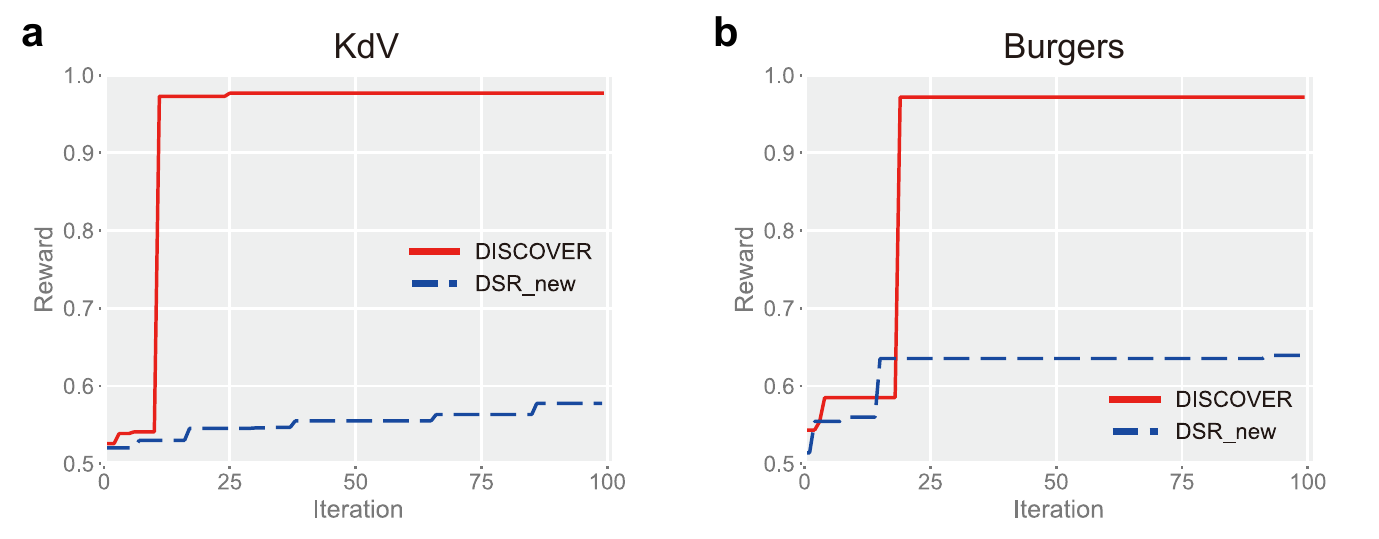}
\caption{\textbf{Maximum reward distribution of DISCOVER and the modified DSR (DSR\_new). }\textbf{a} the KdV equation. \textbf{b} Burgers' equation.}
\label{fig:compare_dsr}
\vspace{-1.0em}
\end{figure}

DSR~\citesupp{dsr} is a powerful framework designed for symbolic regression tasks based on deep reinforcement learning methods. However, compared with our framework, it is insufficient for identifying the governing equations of dynamic systems, especially those described by PDEs. First, the current DSR neither incorporates the computation of partial differential operators, nor is it able to handle sparse and noisy data. Furthermore, even if DSR is modified to be compatible with PDE discovery tasks by expanding the library, introducing regulations and constraints, and the pre-processing part, it is still less efficient and effective to deal with relatively simple dynamic systems. As shown in Fig.~\ref{fig:compare_dsr} a and b, the optimization speed of the modified DSR (DSR\_new) is significantly slower than DISCOVER. After 100 iterations, the maximum reward is still at a low level. This is mainly because the DSR represents coefficients through the constant operator, which expands the search space greatly and is inefficient for determining the form of PDEs. In addition, DSR builds an optimization problem with the total reward as the target and determines the constants by solving it iteratively, which leads to unacceptable computational costs for high-dimensional nonlinear dynamic systems with large data volumes. Taking Burgers' equation as an example, DISCOVER completes 100 time-step iterations in less than one-tenth the time cost by DSR. Therefore, it can be seen that DISCOVER is customized for PDE discovery tasks with better efficiency and scalability.

\subsection{Comparison between structure-aware LSTM and standard LSTM.}
\begin{figure}[htbp]
\includegraphics[width=\linewidth]{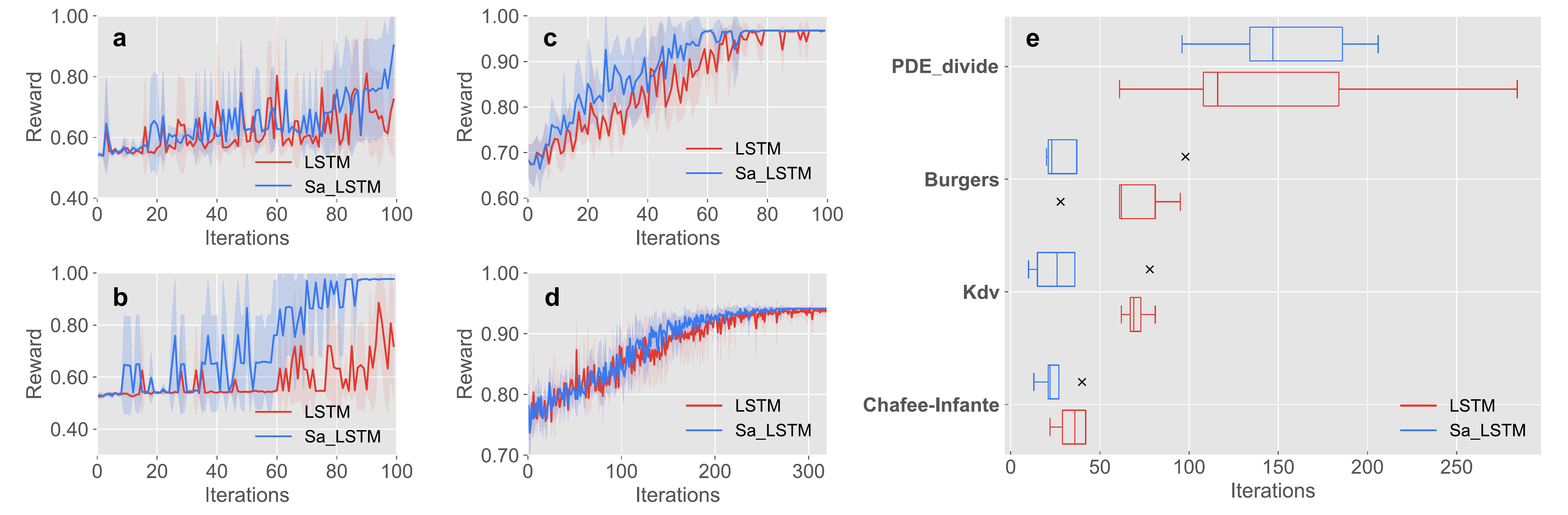}
\caption{\textbf{Maximum reward distribution for four PDEs.} \textbf{a} Burgers' equation.  \textbf{b} The KdV equation. \textbf{c} The Chafee-infante equation. \textbf{d} PDE\_divide. \textbf{e} Iterations needed to discover the correct equation. Sa\_LSTM refers to the proposed structure-aware LSTM agent.}
\label{fig:rnn_box}
\vspace{-1.0em}
\end{figure}

Our model introduces structured information in the agent, so that LSTM can attend to the previously generated tokens and equation structure when predicting the current output. 
However, the standard LSTM agent can only obtain the information from the last token and the composite history information coupled in the memory cell. 
To highlight the rationality and superiority of our model, we compare the performance of the two settings in the above four equations with other hyperparameters being the same. Since the PDE\_compound equation is relatively simple, no specific comparison is made here.  Fig.~\ref{fig:rnn_box} a-d shows the distribution of the rewards under two agent settings during the training process. Since our model can better capture the structural and long-distance information in the equation, the expressions generated in each iteration are closer to the real ones and then lead to larger rewards. Fig.~\ref{fig:rnn_box}e illustrates the number of iterations for discovering the optimal equation form. It can be seen that, except that the PDE\_divide equation is relatively close, the structure-aware LSTM can identify the correct equation form with fewer iterations with almost no increase in the amount of computation.

\subsection{Sensitivity analysis.}
\label{append:ablation}

\begin{figure}[!ht]
\centering
\includegraphics[width=0.8\linewidth]{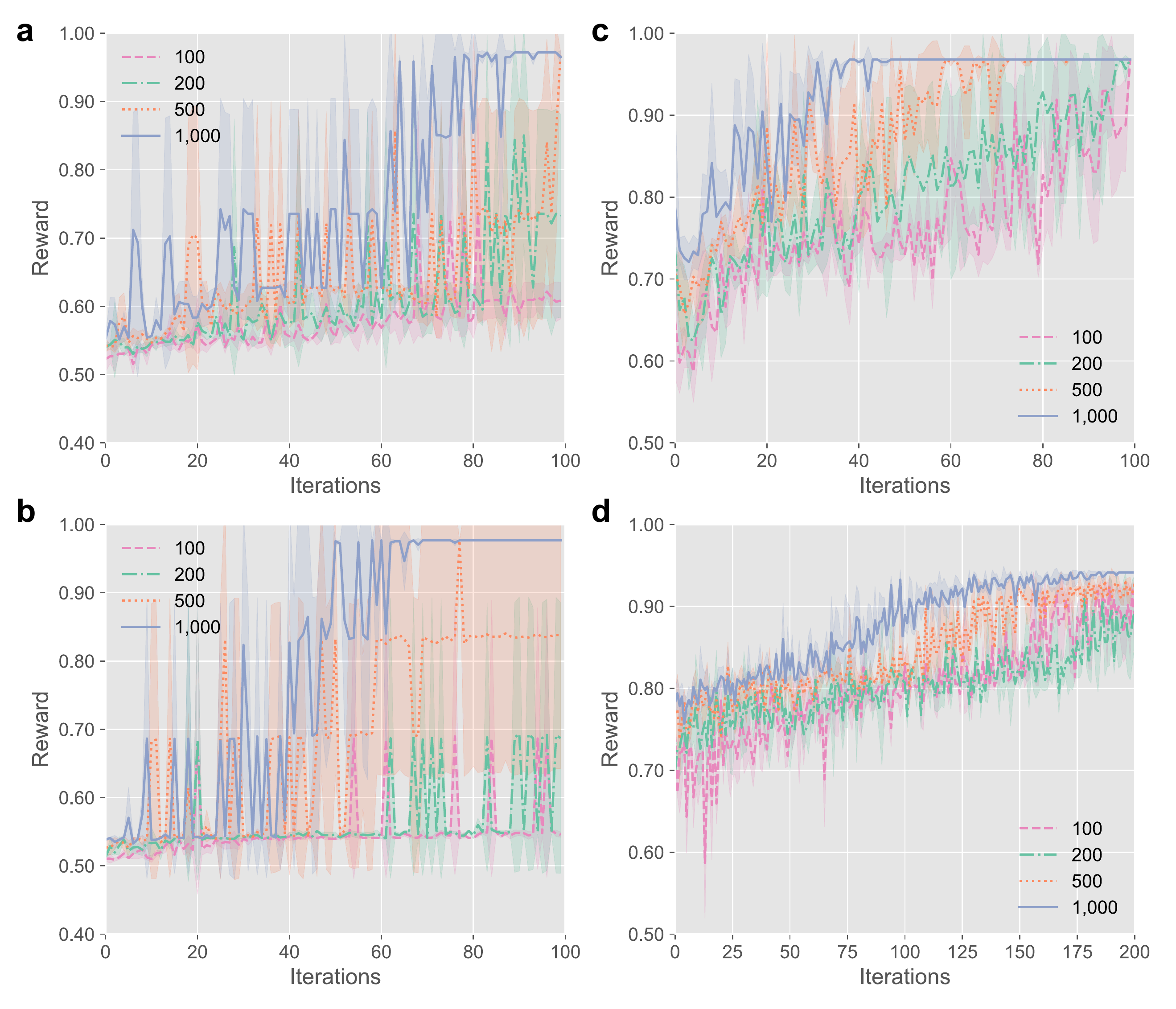}
\caption{\textbf{Maximum reward distribution with different $N$ for four PDEs.} \textbf{a} Burgers' equation. \textbf{b} The KdV equation. \textbf{c} The Chafee-Infante equation. \textbf{d} PDE\_divide.}
\label{fig:batch_num}
\vspace{-1.0em}
\end{figure}
Based on the theory of risk-seeking policy gradient, only $\varepsilon$ fraction of the best expressions is selected to update the agent during the training process to improve the best-case performance. The learning effect and calculation time are directly impacted by the quality and number of the expressions at each iteration. It is mainly affected by two hyperparameters: the total number of generated expressions at each iteration $N$, and the quantile of the rewards $\varepsilon$ used to filter expressions. Then, we focus on discussing their specific impact on the entire optimization process.

\begin{figure}[!ht]
\centering
\includegraphics[width=0.8\linewidth]{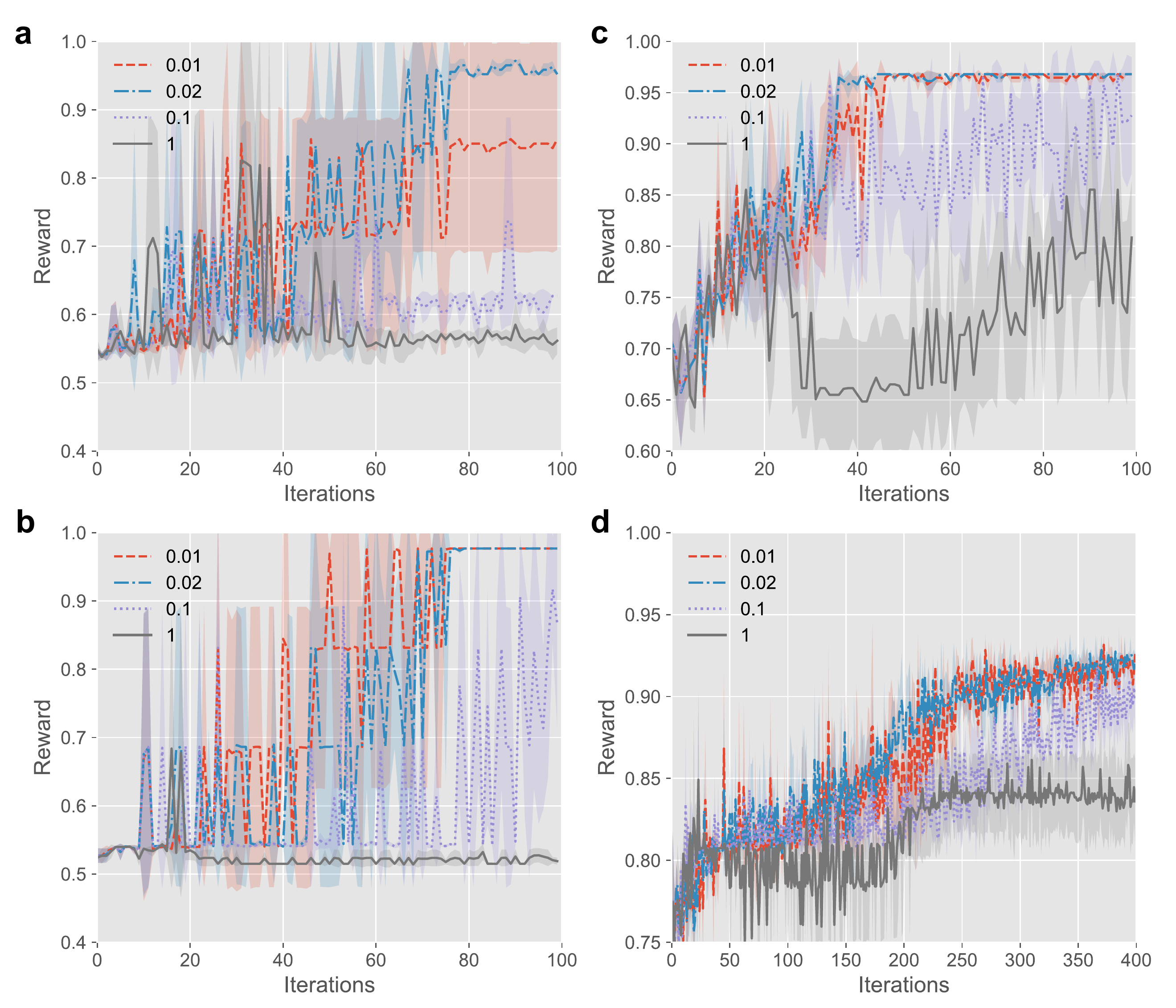}
\caption{ \textbf{Maximum reward distribution with different $\varepsilon$ for four PDEs.} \textbf{a} Burgers' equation.  \textbf{b} The KdV equation. \textbf{c} The Chafee-infante equation. \textbf{d} PDE\_divide.}
\label{fig:e_num}
\vspace{-1.0em}
\end{figure}

Fig.~\ref{fig:batch_num} illustrates the maximum reward distribution of four different PDEs with different $N$. It is obvious that the more expressions are produced in each round, the better the expressions that are ultimately chosen to update the agent, and the fewer iterations it takes to find the optimal equation. However, it is worth noting that generating more expressions also takes more computation resources, and a trade-off needs to be found between the number of generated expressions and the time consumed. In the actual training process, it is necessary to generate as many expressions as possible, especially at the beginning, to avoid becoming stuck in local optima. 
The other parameter $\varepsilon$ determines the proportion of the expressions generated in each round that can actually be used to update the agent. When the parameter is set to 0, all expressions will be used for the update of the agent, which is the standard policy gradient approach. As shown in Fig.~\ref{fig:e_num}, choosing a relatively small and reasonable parameter is necessary for the agent to learn the optimal solution, which can speed-up the training process. However, with the method of policy gradient, each time it is expected that the rewards of all expressions in each batch are maximized, which obviously slows down the update speed, and may even fail to find the final correct expression.

\renewcommand{\dblfloatpagefraction}{.9}

\bibliographysupp{bibliography}
\bibliographystylesupp{naturemag}